%% file: main.tex
\documentclass[twoside,11pt]{article}
\usepackage[hidelinks]{hyperref}
 
\usepackage{jair}
\usepackage{theapa}
\usepackage{rawfonts}
\jairheading{72}{2021}{667-715}{04/2021}{11/2021}
\ShortHeadings{Output Space Entropy Search Framework for Multi-Objective Bayesian Optimization}
{Belakaria, Deshwal, \& Doppa}
\firstpageno{667}

\usepackage{graphicx}
\usepackage{amsthm}
\usepackage{amsmath}
\usepackage{amsfonts} 

\usepackage{float}
\usepackage{algorithm}
\usepackage{algorithmic}
\usepackage{siunitx}

\usepackage{booktabs}       
\usepackage{tcolorbox}
\numberwithin{equation}{section}

\renewcommand{\vec}[1]{\mathbf{#1}}

\begin{document}

\title{Output Space Entropy Search Framework for Multi-Objective Bayesian Optimization}

\author{\name Syrine Belakaria \email syrine.belakaria@wsu.edu \\
      \name Aryan Deshwal \email aryan.deshwal@wsu.edu \\
      \name Janardhan Rao Doppa \email jana.doppa@wsu.edu \\
      \addr School of Electrical Engineering and Computer Science \\ Washington State University\\
      Pullman, Washington 99163, USA}

\maketitle

\begin{abstract}
We consider the problem of black-box multi-objective optimization (MOO) using expensive function evaluations (also referred to as experiments), where the goal is to approximate the true Pareto set of solutions by minimizing the total resource cost of experiments. For example, in hardware design optimization, we need to find the designs that trade-off performance, energy, and area overhead using expensive computational simulations. The key challenge is to select the sequence of experiments to uncover high-quality solutions using minimal resources. In this paper, we propose a general framework for solving MOO problems based on the principle of output space entropy (OSE) search: select the experiment that maximizes the information gained per unit resource cost about the true Pareto front. We appropriately instantiate the principle of OSE search to derive efficient algorithms for the following four MOO problem settings: 1) The most basic {\em single-fidelity setting}, where experiments are expensive and accurate; 2) Handling {\em black-box constraints} which cannot be evaluated without performing experiments; 3) The {\em discrete multi-fidelity setting}, where experiments can vary in the amount of resources consumed and their evaluation accuracy; and 4) The {\em continuous-fidelity setting}, where continuous function approximations result in a huge space of experiments. Experiments on diverse synthetic and real-world benchmarks show that our OSE search based algorithms improve over state-of-the-art methods in terms of both computational-efficiency and accuracy of MOO solutions.

\end{abstract}

\input{Introduction.tex}
\input{genera_background.tex}
\input{RelatedWork.tex}
\input{Mesmo_setup_Approach.tex}
\input{MESMOC_setup_Approach.tex}

\input{MF_OESMO_setup_Approach.tex}
\input{IMOCA_setup_Approach.tex}

\input{Experiments.tex}

\section{Summary and Future Work}

We introduced a novel and general framework for solving multi-objective (MO) Bayesian optimization problems based on the principle of output space entropy (OSE) search. The key idea is to select the sequence of experiments that maximize the information gained per unit cost about the optimal Pareto front. We instantiated this principle appropriately to solve a variety of MO problems from the most basic setting and its constrained version to the multi-fidelity and continuous-fidelity settings. Our comprehensive experimental results on both synthetic and real-world benchmarks showed that all our OSE based algorithms yield consistently better results than state-of-the-art methods, and are more efficient and robust than methods based on input space entropy search. 

Future work includes extending this framework to handle high-dimensional BO problems \cite{BOCK} and combinatorial spaces, e.g., sets, sequences, and graphs \cite{IJCAI-2021,AAAI-2020,COMBO,AAAI-2021,ICML-2021}; and investigating important scientific applications including biological sequence design \cite{yang_protein_review} and molecule design \cite{MOF}.

\vspace{2.0ex}

\noindent {\bf Acknowledgements.} Some of the material in this paper was first published at NeurIPS-2019 \cite{belakaria2019max} and AAAI-2020 \cite{belakaria2020multi}. The authors gratefully acknowledge the support from National Science Foundation (NSF) grants IIS-1845922,  OAC-1910213, and SII-2030159. The views expressed are those of the authors and do not reflect the official policy or position of the NSF.


\appendix
\input{appendix.tex}

\input{main.bbl}
\end{document}

%% file: Introduction.tex

\section{Introduction}


Many engineering and scientific applications involve making design choices to optimize multiple objectives. Some examples include tuning the knobs of a compiler to optimize performance and efficiency of a set of software programs;  designing new materials to optimize strength, elasticity, and durability; and designing hardware to optimize performance, power, and area. There are a few common challenges in solving these kind of multi-objective optimization (MOO) problems: {\bf 1)} The objective functions are unknown and we need to perform expensive experiments to evaluate each candidate design choice, where expense is measured in terms of the consumed resources (physical or computational). For example, performing computational simulations and physical lab experiments for hardware optimization and material design applications respectively. {\bf 2)} The objectives are conflicting in nature and all of them cannot be optimized simultaneously. Therefore, we need to find the {\em Pareto optimal} set of solutions. A solution is called Pareto optimal if it cannot be improved in any of the objectives without compromising some other objective. {\bf 3)} The solutions may need to satisfy black-box constraints, which cannot be evaluated without performing experiments. For example, in aviation power system design applications, we need to find the designs that trade-off total energy and mass while satisfying specific thresholds for motor temperature and voltage of cells. {\bf 4)} We have the ability to perform multi-fidelity experiments (discrete or continuous) to evaluate objective functions via cheaper approximations, which vary in the amount of resources consumed and their accuracy. For example, in hardware design optimization, we can use multi-fidelity simulators for design evaluations. We want to leverage this additional freedom to reduce the overall cost for optimization. Real-world MOO problems come with two or more of the above challenges and the overall goal is to approximate the optimal Pareto set while minimizing the total resource cost of conducted experiments. 

Bayesian Optimization (BO) \cite{shahriari2016taking} is an effective framework to solve black-box optimization problems with expensive function evaluations. The key idea behind BO is to
build a cheap surrogate statistical model, e.g., Gaussian Process \cite{williams2006gaussian}, using the real experimental data; and employ it to intelligently select the sequence of experiments or function evaluations using an acquisition function, e.g., expected improvement (EI) and upper-confidence bound (UCB). There is a large body of literature on single-objective BO algorithms \cite{shahriari2016taking} and their applications including hyper-parameter tuning of machine learning methods \cite{snoek2012practical,kotthoff2017auto}. However, there is relatively less work on the more challenging problem of BO for multiple objective functions (first and second challenges) \cite{PESMO}, very limited work on the constrained multi-objective optimization problem (third challenge), and no prior work on multi-objective optimization in the multi-fidelity setting (fourth challenge). {\em To the best of our knowledge, this is the first work on discrete and continuous-fidelity settings for multi-objective BO within the ML literature} as discussed in the related work section. 

Prior work on multi-objective BO is lacking in the following ways. Many algorithms reduce the problem to single-objective optimization by designing appropriate acquisition functions, e.g., expected improvement in Pareto hypervolume \cite{knowles2006parego,emmerich2008computation}. This can potentially lead to aggressive exploitation behavior. Additionally, algorithms to optimize Pareto Hypervolume (PHV) based acquisition functions scale poorly as the number of objectives and the dimensionality of input space grows. There are also methods that rely on {\em input space entropy} based acquisition function \cite{PESMO} to select the candidate inputs for evaluation. However, it is computationally expensive to approximate and optimize this acquisition function.

In this paper, we study a general framework for solving a large-class of black-box MOO problems based on the principle of {\em output space entropy (OSE)} search \cite{MES,hoffman2015output}. Our work is inspired by the prior success of the OSE principle for solving single-objective BO problems and is an extension of \citeA{MES} to several multi-objective optimization settings. The key idea is to select the input and fidelity vector (if applicable) that maximizes the information gain per unit resource cost about the optimal Pareto front in each iteration.  Output space entropy search has many advantages over algorithms based on input space entropy search \cite{belakaria2019max}: a) it allows much tighter approximation; b) it is cheaper to compute; and c) it naturally lends itself to robust optimization with respect to the number of samples used for acquisition function computation.  We appropriately instantiate the OSE principle to derive efficient algorithms for solving four qualitatively different MOO problems: the most basic single-fidelity setting \cite{belakaria2019max}, MOO with black-box constraints, discrete multi-fidelity  setting \cite{belakaria2020multi}, and continuous-fidelity setting. Comprehensive experiments on diverse synthetic and real-world benchmarks show that our OSE search based algorithms are computationally-efficient and perform better than the state-of-the-art algorithms.

\vspace{1.0ex}

\noindent {\bf Contributions.} The main contribution of this paper is the development and evaluation of multi-objective BO algorithms based on the principle of output space entropy search for four different MOO problem settings. Specific contributions include the following:
\begin{itemize}
\setlength\itemsep{0em} 
\item Development of an approach referred to as MESMO to solve the most basic MOO problem in the single-fidelity setting, where experiments are expensive and accurate \cite{belakaria2019max}.
\item Development of an approach referred to as MESMOC to handle MOO problems with black-box constraints, which cannot be evaluated without performing experiments.
\item Development of an approach referred to as MF-OSEMO to solve MOO problems in the discrete multi-fidelity setting, where experiments can vary in the amount of resources consumed and their evaluation accuracy \cite{belakaria2020multi}.
\item Development of an approach referred to as iMOCA to solve MOO problems in the continuous-fidelity setting, where continuous function approximations result in a huge space of experiments with varying cost. We provide two qualitatively different approximations for iMOCA.
\item Experimental evaluation on diverse synthetic and real-world benchmark problems to demonstrate the effectiveness of the proposed algorithms over existing  MOO algorithms and a naive continuous-fidelity baseline.
\item Open-source code for all methods: MESMO\footnote{github.com/belakaria/MESMO}, MESMOC\footnote{github.com/belakaria/MESMOC},
MF-OSEMO\footnote{github.com/belakaria/MF-OSEMO}, and
iMOCA\footnote{github.com/belakaria/iMOCA}

\end{itemize}

%% file: genera_background.tex
\section{Background and Problem Setup}

In this section, we first provide an overview of the generic Bayesian optimization framework. Next, we formally define the different MOO problem settings considered in this work.

\subsection{Bayesian Optimization Framework}

Bayesian Optimization (BO) is a very efficient framework to solve global optimization problems using {\em black-box evaluations of expensive objective functions}. Let $\mathfrak{X} \subseteq \Re^d$ be an input space. In the single-objective BO formulation, we are given an unknown real-valued objective function $f: \mathfrak{X} \mapsto \Re$, which can evaluate each input $\vec{x} \in \mathfrak{X}$ to produce an evaluation $y$ = $f(\vec{x})$.  Each evaluation $f(\vec{x})$ is expensive in terms of the consumed resources. The main goal is to find an input $\vec{x^*} \in \mathfrak{X}$ that approximately optimizes $f$ by performing a limited number of function evaluations. BO algorithms learn a cheap surrogate model from training data obtained from past function evaluations. They intelligently select the next input for evaluation by trading-off exploration and exploitation to quickly direct the search towards optimal inputs. The three key elements of BO framework are:

\vspace{1.0ex}

\hspace{2.0ex} {\bf 1) Statistical Model} of the true function $f(x)$. {\em Gaussian Process (GP)} \cite{williams2006gaussian} is the most commonly used model. A GP over a space $\mathfrak{X}$ is a random process from $\mathfrak{X}$ to $\Re$. It is characterized by a mean function $\mu : \mathfrak{X} \mapsto \Re$ and a covariance or kernel function $\kappa : \mathfrak{X} \times \mathfrak{X} \mapsto \Re$. If a function $f$ is sampled from $\mathcal{GP}(\mu, \kappa)$, then $f(x)$ is distributed normally $\mathcal{N}(\mu(x), \kappa(x,x))$ for a set of inputs from $x \in \mathcal{X}$.

\vspace{1.0ex}

\hspace{2.0ex} {\bf 2) Acquisition Function} ($\alpha$) to score the utility of evaluating a candidate input $\vec{x} \in \mathfrak{X}$ based on the statistical model. Some popular acquisition functions in the single-objective literature include expected improvement (EI), upper confidence bound (UCB), predictive entropy search (PES) \cite{PES}, and max-value entropy search (MES) \cite{MES}. 

\vspace{1.0ex}

\hspace{2.0ex} {\bf 3) Optimization Procedure} to select the best scoring candidate input according to $\alpha$ depending on statistical model. DIRECT \cite{jones1993lipschitzian} is a very popular approach for acquisition function optimization.
\begin{figure}[t]
    \centering
    \includegraphics[width=0.9\columnwidth]{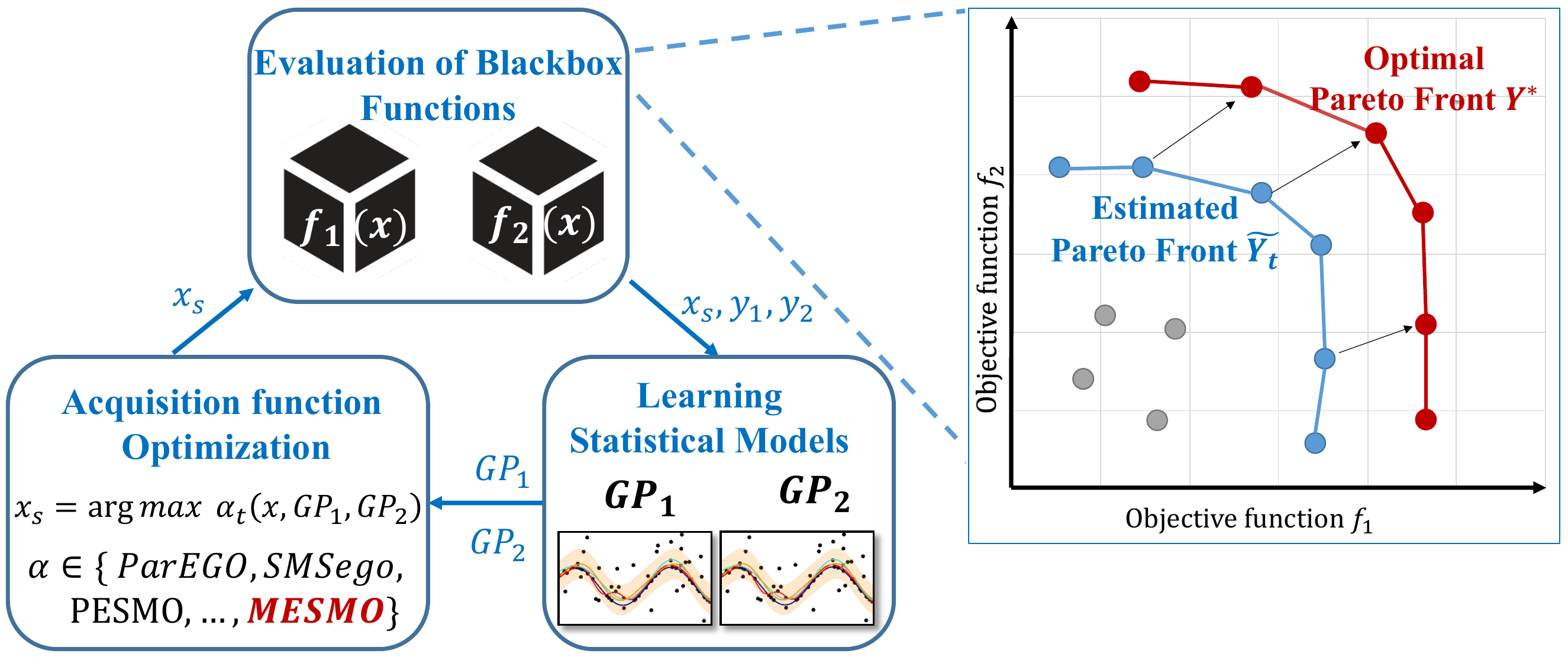}
    \caption{Overview of the Bayesian optimization process for two objective functions ($k$=2). }
    \label{fig:BO}
\end{figure}

\subsection{Multi-Objective Optimization Problem Setting Overview }   

Multi-objective optimization (MOO) problems can be formalized in terms of the following key elements: number of objectives, necessity to satisfy black-box constraints, and availability of cheaper approximations or fidelities (discrete/continuous) for function evaluations. Below we provide a brief overview of the four different MOO problem settings that are addressed in this paper noting that a more detailed problem setup is specified under each technical section.

\vspace{1.0ex}

\noindent\textbf{Basic Multi-objective Optimization Problem.} The goal is to maximize real-valued objective functions $f_1(\vec{x}), f_2(\vec{x}),\cdots,f_K(\vec{x})$, with  $K \geq 2$, over continuous space  $\mathfrak{X} \subseteq \Re^d$. Each evaluation (also called an experiment) of an input $\vec{x}\in \mathfrak{X}$ produces a vector of objective values $\vec{y}$ = $(y_1, y_2,\cdots,y_K)$ where $y_i = f_i(x)$ for all $i \in \{1,2, \cdots, K\}$.

\vspace{1.0ex}

\noindent\textbf{MOO Problem with Constraints.} This is a generalization of the basic MOO problem, where we need to satisfy some black-box constraints. Our goal is to maximize real-valued objective functions $f_1(\vec{x}), f_2(\vec{x}),\cdots,f_K(\vec{x})$, with  $K \geq 2$,  while satisfying $L$ black-box constraints of the form $C_1(x)\geq 0, C_2(x)\geq 0,\cdots,C_L(x)\geq 0$ over continuous space  $\mathfrak{X} \subseteq \Re^d$. Each evaluation of an input $\vec{x}\in \mathfrak{X}$ produces a vector of objective values and constraint values $\vec{y}$ = $(y_{f_1}, y_{f_2},\cdots,y_{f_K},y_{c_1} \cdots y_{c_L})$ where $y_{f_j} = f_j(x)$ for all $j \in \{1,2, \cdots, K\}$ and $y_{c_i} = C_i(x)$ for all $i \in \{1,2, \cdots, L\}$.

\vspace{1.0ex}

\noindent\textbf{MOO Problem with Discrete Multi-fidelity Experiments.} This is a general version of the MOO problem, where we have access to $M_j$ fidelities for each function $f_j$ that vary in the amount of resources consumed and the accuracy of evaluation. The evaluation of an input $\vec{x}\in \mathfrak{X}$ with fidelity vector $\vec{m} = [m_1, m_2, \cdots, m_K]$ produces an evaluation vector of $K$ values denoted by $\vec{y}^{\vec{m}} \equiv [y_1^{(m_1)}, \cdots, y_K^{(m_K)}]$, where $y_j^{(m_j)} = f_j^{(m_j)}(x)$ for all $j \in \{1,2, \cdots, K\}$. 

\vspace{1.0ex}

\noindent\textbf{MOO Problem with Continuous-fidelity Experiments.} In this general version of the multi-fidelity setting, we have access to $g_i(\vec{x},z_i)$ where $g_i$ is an alternative function through which we can evaluate cheaper approximations of $f_i$ by varying the fidelity variable $z_i \in \mathcal{Z}$ (continuous function approximations). The evaluation of an input $\vec{x}\in \mathcal{X}$ with fidelity vector $\vec{z} = [z_1, z_2, \cdots, z_K]$ produces an evaluation vector of $K$ values denoted by $\vec{y} \equiv [y_1, y_2,\cdots, y_K]$, where $y_i = g_i(\vec{x},z_i)$ for all $i \in \{1,2, \cdots, K\}$.

\begin{table*}[t]
    \centering
    \resizebox{0.95\linewidth}{!}{
    \begin{tabular}{|c|c|}
    \hline
    {\bf Notation} & {\bf Definition} \\
    \hline \hline
       $\vec{x}, \vec{y}, \vec{f}, \vec{m}$ & bold notation represents vectors\\
       \hline
        $\vec{x}$ & input vector of $d$ dimensions \\
       \hline
       
       $[n]$ & set of first $n$ natural numbers $\{1,2,\cdots, n\}$ \\
       \hline
       $f_1, f_2, \cdots, f_K$  & true objective functions \\
        \hline
       $C_1, C_2, \cdots, C_L$  & Constraints functions \\
       \hline
       $\Tilde{f}_j$ & function sampled from the highest fidelity of the $j$th Gaussian process model\\
\hline
       $\mathcal{X}$& Input space \\
\hline 
       $I$ & Information gain \\

       \hline
      
$\mathcal{Y}^*$ & true pareto front of the objective functions $[f_1, f_2, \cdots, f_K]$ \\      \hline 
$\mathcal{Y}_s^*$ &  Pareto front of the sampled functions $[\Tilde{f}_1, \Tilde{f}_2, \cdots, \Tilde{f}_K]$\\ 
\hline 
    \end{tabular}}
    \caption{Table describing the general mathematical notations.}
    \label{table:notations}
\end{table*}

%% file: RelatedWork.tex

\section{Related Work}

In this section, we discuss prior work from the BO literature that is related to the four MOO problem settings considered in this paper.
\vspace{2.0ex}

\noindent {\bf Single-fidelity Multi-Objective Optimization.} 
There is a family of model based multi-objective BO algorithms that reduce the problem to single-objective optimization. The ParEGO method \cite{knowles2006parego} employs random scalarization for this purpose: scalar weights of $K$ objective functions are sampled from a uniform distribution to construct a single-objective function and expected improvement is employed as the acquisition function to select the next input for evaluation. ParEGO is simple and fast, but more advanced approaches often outperform it. Many methods optimize the Pareto hypervolume (PHV) metric \cite{emmerich2008computation} that captures the quality of a candidate Pareto set. This is done by extending the standard acquisition functions to PHV objective, e.g., expected improvement in PHV (EHI) \cite{emmerich2008computation} and probability of improvement in PHV (SUR) \cite{picheny2015multiobjective}. Unfortunately, algorithms to optimize PHV based acquisition functions scale very poorly and are not feasible for more than two objectives. SMSego is a relatively faster method \cite{ponweiser2008multiobjective}. To improve scalability, the gain in hypervolume is computed over a limited set of points: SMSego finds those set of points by optimizing the posterior means of the GPs. A common drawback of this family of algorithms is that reduction to single-objective optimization can potentially lead to more exploitation behavior resulting in sub-optimal solutions.

PAL \cite{zuluaga2013active}, PESMO \cite{PESMO}, and the concurrent works USeMO \cite{Usemo} and MESMO \cite{belakaria2019max} are principled algorithms based on information theory. PAL tries to classify the input points based on the learned models into three categories: Pareto optimal, non-Pareto optimal, and uncertain. In each iteration, it selects the candidate input for evaluation towards the goal of minimizing the size of uncertain set. PAL provides theoretical guarantees, but it is only applicable for input space $\mathfrak{X}$ with finite set of discrete points. USeMO is a general framework that iteratively generates a cheap Pareto front using the surrogate models and then selects the input with highest uncertainty for evaluation. PESMO \cite{PESMO} relies on input space entropy based acquisition function and  iteratively selects the input that maximizes the information gained about the optimal Pareto set $\mathcal{X}^*$. Unfortunately, optimizing this acquisition function poses significant challenges: a) it requires a series of approximations, which can be potentially sub-optimal; b) the optimization, even after approximations, is expensive; and c) the performance is strongly dependent on the number of Monte-Carlo samples. In comparison, our proposed output space entropy based acquisition function partially overcomes the above challenges, and allows efficient and robust optimization with respect to the number of samples used for acquisition function computation. More specifically, the time complexities of acquisition function computation in PESMO and MESMO ignoring the time to solve the cheap MO problem that is common for both algorithms are $\mathcal{O}(SKm^3)$ and $\mathcal{O}(SK)$ respectively, where $S$ is the number of Monte-Carlo samples, $K$ is the number of objectives, and $m$ is the size of the sample Pareto set in PESMO. In fact, PESMO formulation relies on an expensive and high-dimensional ($l\cdot d$ dimensions) distribution over the input space, where $l$ is size of the optimal Pareto set $\mathcal{X}^*$ while MESMO relies on a computationally cheap and low-dimensional distribution over the output space ($l\cdot K$ dimensions, which is considerably less than $l \cdot $d as $K \ll d$ in practice). Additionally, \citeA{belakaria2019max} demonstrated  that MESMO is very robust and performs very well even with one sample.
\vspace{2.0ex}

\noindent {\bf Constrained Multi-Objective Optimization.} 
There exists very limited prior work to address constrained MO problems \cite{garrido2019predictive,feliot2017bayesian}. PESMOC \cite{garrido2019predictive} is the current state-of-the-art method for this problem setting. PESMOC extends the information-theoretic approach PESMO that relies on the principle of input space entropy search to the constrained setting. As a consequence, it inherits the drawbacks of PESMO. Our proposed MESMOC algorithm based on OSE search is intended to improve over PESMOC. MESMOC+ \cite{fernandez2020max} is a concurrent work that also employs the principle of output space entropy search to solve constrained multi-objective optimization problems. However, this paper uses a completely different approximation of the information gain leading to a different expression of the acquisition function. This method employs a series of complex mathematical approximations based on Assumed Density Filtering (ADF). Our proposed MESMOC algorithm uses the truncated Gaussian distribution approximation that results in a closed-form expression, fast, and easy to implement acquisition function. Additionally, the ADF based method \cite{fernandez2020max} considers blackbox constraints only in the acquisition function definition while MESMOC addresses the constraints both in the acquisition function expression and in the acquisition function optimization to ensure the selection of valid inputs. 
\vspace{2.0ex}

\noindent {\bf Multi-fidelity Single-Objective Optimization.} 
Acquisition functions (AFs) for single-fidelity and single-objective BO are extensively studied \cite{shahriari2016taking}. AFs can be broadly classified into two categories. First, {\em myopic} AFs rely on improving a ``local'' measure of utility (e.g., expected improvement). Second, {\em non-myopic} AFs measure the ``global'' utility of evaluating a candidate input for solving the black-box optimization problem (e.g., predictive entropy search). Canonical examples of myopic acquisition function include expected improvement (EI) and upper-confidence bound (UCB). EI was extended to multi-fidelity setting \cite{huang2006sequential,picheny2013quantile,lam2015multifidelity}. The popular GP-UCB method \cite{gp-ucb} was also extended to multi-fidelity setting with discrete fidelities \cite{kandasamy2016gaussian} and continuous fidelities \cite{kandasamy2017multi}. 
Entropy based methods fall under the category of non-myopic AFs. Some examples include entropy search (ES) \cite{entropy_search} and predictive entropy search (PES) \cite{PES}. Their multi-fidelity extensions include MT-ES \cite{swersky2013multi,klein2017fast} and MF-PES \cite{zhang2017information,mcleod2017practical}. Unfortunately, they inherit the computational difficulties of the original ES and PES. Max-value entropy search (MES) \cite{MES} and output space predictive entropy search \cite{hoffman2015output} are recent approaches that rely on the principle of output space entropy (OSE) search. Prior work \cite{MES} has shown advantages of OSE search in terms of compute-time, robustness, and accuracy over input space entropy search methods.
Recent work \cite{song2018general} proposed a general approach based on mutual information. \citeA{takeno2019multi} extended MES to multi-fidelity setting and showed its effectiveness over MF-PES. MUMBO \cite{moss2020mumbo} extended MES to the continuous-fidelity and multi-task setting. 

\vspace{2.0ex}

\noindent {\bf Multi-fidelity Multi-Objective Optimization.} 
Prior work outside ML literature has considered domain-specific methods that employ single-fidelity multi-objective approaches in the context of multi-fidelity setting by using the lower fidelities {\em only as an initialization} \cite{kontogiannis2018comparison,ariyarit2017multi}. Specifically, \citeA{ariyarit2017multi} employs the single-fidelity algorithm based on expected hypervolume improvement acquisition function and \citeA{kontogiannis2018comparison} employs an algorithm that is very similar to SMSego. Also, both these methods model all fidelities with the same GP and assume that higher fidelity evaluation is a sum of lower-fidelity evaluation and offset error. These are strong assumptions and may not hold in general multi-fidelity settings including the problems from our experimental evaluation. Our proposed MF-OSEMO \cite{belakaria2020multi} and iMOCA algorithms (generalized versions of MESMO \cite{belakaria2019max} solve MOO problem in discrete and continuous-fidelity settings respectively using the principle of output space entropy search and leverage some technical ideas from the prior work on single-objective optimization. We are not aware of any prior work on generic discrete/continuous-fidelity algorithms for MOO problems in the BO literature.

%% file: Mesmo_setup_Approach.tex


\section{MESMO Algorithm for the Basic MOO Problem}
\label{Section-MESMO}
In this section, we address the most basic MOO problem in the single-fidelity setting, where the goal is to optimize multiple black-box objective functions. To solve this problem, we propose an algorithm referred to as {\em {\bf M}ax-value {\bf E}ntropy {\bf S}earch for {\bf M}ulti-objective {\bf O}ptimization} (MESMO). In what follows, we first describe the problem setup and surrogate models. Next, we mathematically describe the output space entropy based acquisition function and provide an algorithmic approach to efficiently compute it. 

\vspace{1.0ex} 

\noindent
{\bf Problem Setup (Basic Multi-Objective Optimization Problem).}
The goal is to maximize real-valued objective functions $f_1(\vec{x}), f_2(\vec{x}),\cdots,f_K(\vec{x})$, with  $K \geq 2$, over continuous space  $\mathfrak{X} \subseteq \Re^d$. Each evaluation (also called an experiment) of an input $\vec{x}\in \mathfrak{X}$ produces a vector of objective values $\vec{y}$ = $(y_1, y_2,\cdots,y_K)$ where $y_i = f_i(x)$ for all $i \in \{1,2, \cdots, K\}$. We say that an point $\vec{x}$ {\em Pareto-dominates} another point $\vec{x'}$ if $f_i(\vec{x}) \geq f_i(\vec{x'}) \hspace{1mm} \forall{i}$ and there exists some $j \in \{1, 2, \cdots,K\}$ such that $f_j(\vec{x}) > f_j(\vec{x'})$. The optimal solution of MOO problem is a set of points $\mathcal{X}^* \subset \mathfrak{X}$ such that no point $\vec{x'} \in \mathfrak{X} \setminus \mathcal{X}^*$ Pareto-dominates a point $\vec{x} \in \mathcal{X}^*$. The solution set $\mathcal{X}^*$ is called the optimal {\em Pareto set} and the corresponding set of function values $\mathcal{Y}^*$  is called the optimal {\em Pareto front}. The goal of multi-objective BO is to approximate $\mathcal{X}^*$ while minimizing the number of expensive function evaluations. In the application of hardware design optimization, $\vec{x}\in \mathfrak{X}$ is a candidate hardware design; evaluation of of design $\vec{x}$ to get output objectives such as power, performance, and area involve performing computationally-expensive simulation to mimic the real hardware; and our goal is to find the optimal Pareto set of hardware designs to trade-off power, performance, and area. Table~\ref{table:notations} contains all the mathematical notations used in this section.

\vspace{1.0ex} 

\noindent{\bf Surrogate Models.} Gaussian processes (GPs) are shown to be effective surrogate models in prior work on single and multi-objective BO \cite{PES,EST,MES,gp-ucb,PESMO}. Similar to prior work \cite{PESMO}, we model the objective functions $f_1, f_2,\cdots,f_K$ using $K$ independent GP models $\mathcal{GP}_1,\mathcal{GP}_2,\cdots,\mathcal{GP}_K$ with zero mean and i.i.d. observation noise. Let $\mathcal{D} = \{(\vec{x}_i, \vec{y}_i)\}_{i=1}^{t-1}$ be the training data from past $t{-1}$ function evaluations, where 
$\vec{x}_i \in \mathfrak{X}$ is an input and $\vec{y}_i = \{y^i_1,y^i_2,\cdots,y^i_K\}$ is the output vector resulting from evaluating functions $f_1, f_2,\cdots,f_K$ at $\vec{x}_i$. We learn surrogate models $\mathcal{GP}_1,\mathcal{GP}_2,\cdots,\mathcal{GP}_K$ from $\mathcal{D}$.
\vspace{1.0ex} 
\subsection{MESMO Algorithm}
\noindent{\bf Output Space Entropy Based Acquisition Function.} Input space entropy based methods like PESMO \cite{PESMO} selects the next candidate input $\vec{x}_{t}$ (for ease of notation, we drop the subscript in below discussion) by maximizing the information gain about the optimal Pareto set $\mathcal{X}^*$. The acquisition function based on input space entropy is given as follows:
\begin{align}
 \alpha(\vec{x}) &= I(\{\vec{x}, \vec{y}\}, \mathcal{X}^* \mid D) \label{eqn_orig_inf_gain}\\ 
 &= H(\mathcal{X}^* \mid D) - \mathbb{E}_y [H(\mathcal{X}^* \mid D \cup \{\vec{x}, \vec{y}\})] \label{eqn_exp_redn} \\
 &= H(\vec{y} \mid D, \vec{x}) - \mathbb{E}_{\mathcal{X}^*} [H(\vec{y} \mid D, \vec{x}, \mathcal{X}^*)] \label{eqn_symmetric_pesmo}
\end{align} 

Information gain is defined as the expected reduction in entropy $H(.)$\footnote{The conditioning on $D$ and $\vec{x}$ in $H(\vec{y} \mid D, \vec{x})$ is on fixed values and not random variables} of the posterior distribution $P(\mathcal{X}^* \mid D)$ over the optimal Pareto set $\mathcal{X}^*$ as given in equations (\ref{eqn_exp_redn}) and (\ref{eqn_symmetric_pesmo}) (resulting from symmetric property of information gain). This mathematical formulation relies on an expensive and high-dimensional ($l\cdot d$ dimensions) distribution $P(\mathcal{X}^* \mid D)$, where $l$ is size of the optimal Pareto set $\mathcal{X}^*$. Furthermore, optimizing the second term in r.h.s poses significant challenges: a) it requires a series of approximations \cite{PESMO} which can be potentially sub-optimal; and b) the optimization, even after approximations, is expensive c) the performance is strongly dependent on the number of Monte-Carlo samples.

To overcome the above challenges of computing input space entropy based acquisition function, we take an alternative route and propose to maximize the information gain about the optimal {\bf Pareto front} $\mathcal{Y}^*$. This is equivalent to expected reduction in entropy over the Pareto front $\mathcal{Y}^*$, which relies on a computationally cheap and low-dimensional ($l \cdot K$ dimensions, which is considerably less than $l \cdot d$ as $K \ll d$ in practice) distribution $P( \mathcal{Y}^* \mid D)$. Our acquisition function that maximizes the information gain between the next candidate input for evaluation $\vec{x}$ and Pareto front $\mathcal{Y}^*$ is given as:
\begin{align}
 \alpha(\vec{x}) &= I(\{\vec{x}, \vec{y}\}, \mathcal{Y}^* \mid D) \\ 
 &= H(\mathcal{Y}^* \mid D) - \mathbb{E}_y [H(\mathcal{Y}^* \mid D \cup \{\vec{x}, \vec{y}\})] \\
 &= H(\vec{y} \mid D, \vec{x}) - \mathbb{E}_{\mathcal{Y}^*} [H(\vec{y} \mid D, \vec{x}, \mathcal{Y}^*)] \label{eqn_symmetric_mesmo}
\end{align}

The first term in the r.h.s of equation (\ref{eqn_symmetric_mesmo}) (entropy of a factorizable K-dimensional  Gaussian distribution $P(\vec{y}\mid D, \vec{x}$)) can be computed in closed form as shown below:
\begin{align}
 H(\vec{y} \mid D, \vec{x}) = \frac{K(1+\ln(2\pi))}{2} + \sum_{j = 1}^K \ln (\sigma_j(\vec{x})) \label{eqn_unconditioned_entropy}
\end{align}
where $\sigma_i^2(\vec{x})$ is the predictive variance of $i^{th}$ GP at input $\vec{x}$. The second term in the r.h.s of equation (\ref{eqn_symmetric_mesmo}) is an expectation over the Pareto front $\mathcal{Y}^*$. We can approximately compute this term via Monte-Carlo sampling as shown below: 
\begin{align}
 \mathbb{E}_{\mathcal{Y}^*} [H(\vec{y} \mid D, \vec{x}, \mathcal{Y}^*)] \simeq \frac{1}{S} \sum_{s = 1}^S [H(\vec{y} \mid D, \vec{x}, \mathcal{Y}^*_s)] \label{eqn_summation}
\end{align}
where $S$ is the number of samples and $\mathcal{Y}^*_s$ denote a sample Pareto front. The main advantages of our acquisition function are: computational efficiency and robustness to the number of samples. Our experiments demonstrate these advantages over input space entropy based acquisition function. 

There are two key algorithmic steps to compute equation (\ref{eqn_summation}). We want to know: 1) how to compute Pareto front samples $\mathcal{Y}^*_s$?; and 2) and how to compute the entropy with respect to a given Pareto front sample $\mathcal{Y}^*_s$? We provide solutions for these two questions.

\vspace{1.0ex}

\hspace{2.0ex} {\bf 1) Computing Pareto Front Samples via Cheap Multi-Objective optimization.} To compute a Pareto front sample $\mathcal{Y}^*_s$, we first sample functions from the posterior GP models via random Fourier features \cite{PES,random_fourier_features} and then solve a cheap multi-objective optimization over the $K$ sampled functions.

\hspace{3.5ex} {\em Sampling functions from posterior GP.} Similar to prior work \cite{PES,PESMO,MES}, we employ random Fourier features based sampling procedure. We approximate each GP prior as $\Tilde{f} = \phi(\vec{x})^T \theta$, where $\theta \sim N(0, \vec{I})$. The key idea behind random Fourier features is to construct each function sample $\Tilde{f}(\vec{x})$ as a finitely parametrized approximation: $\phi(\vec{x})^T \theta$, where $\theta$ is sampled from its corresponding posterior distribution conditioned on the data $\mathcal{D}$ obtained from past function evaluations: $\theta | \mathcal{D} \sim N(\vec{A^{-1}\Phi^Ty}_n, \sigma^2\vec{A^{-1}})$, where $\vec{A} = \vec{\Phi^T\Phi} + \sigma^2 \vec{I}$ and $\Phi^T = [\phi(\vec{x}_1),\cdots,\phi(\vec{x}_{t-1})]$.

\hspace{3.5ex}{\em Cheap MO solver.} We sample $\Tilde{f}_i$ from GP model $\mathcal{GP}_i$ for each of the $K$ functions as described above. A {\em cheap} multi-objective optimization problem over the $K$ sampled functions $\Tilde{f}_1,\Tilde{f}_2,\cdots,\Tilde{f}_k$ is solved to compute sample Pareto front $\mathcal{Y}^*_s$. This cheap multi-objective optimization also allows us to capture the interactions between different objectives. We employ the popular NSGA-II algorithm \cite{deb2002nsga} to solve the MO problem with cheap objective functions noting that any other algorithm can be used to similar effect. 

\vspace{1.0ex}

\hspace{2.0ex}{\bf 2) Entropy Computation with a Sample Pareto Front.}
Let $\mathcal{Y}^*_s = \{\vec{v}^1, \cdots, \vec{v}^l \}$ be the sample Pareto front, where $l$ is the size of the Pareto front and each $\vec{v}^i = \{v_1^i,\cdots,v_K^i\}$ is a $K$-vector evaluated at the $K$ sampled functions. The following inequality holds for each component $y_j$ of the $K$-vector $\vec{y} = \{y_1, \cdots, y_K\}$ in the entropy term $H(\vec{y} \mid D, \vec{x}, \mathcal{Y}^*_s)$:
\begin{align}
 y_j &\leq y_{j_s}^{*} \quad \forall j \in \{1,\cdots,K\} \label{inequality}
\end{align}
where $y_{j_s}^{*} =  \max \{v^1_j, \cdots v^l_j \}$. The inequality essentially says that the $j^{th}$ component of $\vec{y}$ (i.e., $y_j$) is upper-bounded by a value obtained by taking the maximum of $j^{th}$ components of all $l$ $K$-vectors in the Pareto front $\mathcal{Y}^*_s$. This inequality can be proven by a contradiction argument. Suppose there exists some component $y_j$ of $\vec{y}$ such that $ y_j > y_{j_s}^{*}$. However, by definition, $\vec{y}$ is a non-dominated point because no point dominates it in the $j$th dimension. This results in $\vec{y} \in \mathcal{Y}^*_s$, which is a contradiction. Therefore, our hypothesis that $ y_j > y_{j_s}^{*}$ is incorrect and inequality (\ref{inequality}) holds.

By combining the inequality (\ref{inequality}) and the fact that each function is modeled as a GP, we can approximate each component $y_j$ as a truncated  Gaussian distribution since the distribution of $y_j$ needs to satisfy $ y_j \leq y_{j_s}^{*}$. Furthermore, a common property of entropy measure allows us to decompose the entropy of a set of independent variables into a sum over entropies of individual variables \cite{information_theory}:
\begin{align}
H(\vec{y} \mid D, \vec{x}, \mathcal{Y}^*_s) = \sum_{j=1}^K H(y_j|D, \vec{x}, y_{j_s}^{*}) \label{eqn_sep_ineq}
\end{align}

The r.h.s is a summation over entropies of $K$ variables $\{y_1, \cdots, y_K\}$. The probability distribution of each variable $y_j$ is a truncated  Gaussian with upper bound $y_{j_s}^{*}$ \cite{entropy_handbook}. The differential entropy for each $y_j$ is given as:
\begin{align}
 H(y_j \mid D, \vec{x}, \mathcal{Y}^*_s) \simeq \left[\frac{(1 + \ln(2\pi))}{2}+ \ln(\sigma_j(\vec{x})) + \ln \Phi(\gamma_s^j(\vec{x})) - \frac{\gamma_s^j(\vec{x}) \phi(\gamma_s^j(\vec{x}))}{2\Phi(\gamma_s^j(\vec{x}))}\right] \label{eqn_entropy_f_j}
\end{align}

equation (\ref{eqn_sep_ineq}) and equation (\ref{eqn_entropy_f_j}) give the followong expression of $H(\vec{y} \mid D, \vec{x}, \mathcal{Y}^*_s)$.
\begin{align}
 H(\vec{y} \mid D, \vec{x}, \mathcal{Y}^*_s) \simeq \sum_{j=1}^K \left[\frac{(1 + \ln(2\pi))}{2}+ \ln(\sigma_j(\vec{x})) + \ln \Phi(\gamma_s^j(\vec{x})) - \frac{\gamma_s^j(\vec{x}) \phi(\gamma_s^j(\vec{x}))}{2\Phi(\gamma_s^j(\vec{x}))}\right]
 \label{eqn_entropy_closed}
\end{align}

where $\gamma_s^j(x) = \frac{y_{j_s}^{*} - \mu_j(\vec{x})}{\sigma_j(\vec{x})}$, and $\phi$ and $\Phi$ are the p.d.f and c.d.f of a standard normal distribution respectively. By combining equations (\ref{eqn_unconditioned_entropy}) and (\ref{eqn_entropy_closed}) with equation (\ref{eqn_symmetric_mesmo}), we get the final form of our acquisition function as shown below:
\begin{align}
\alpha(\vec{x}) \simeq \frac{1}{S} \sum_{s=1}^S \sum_{j=1}^K \left[ \frac{\gamma_s^j(\vec{x}) \phi(\gamma_s^j(\vec{x}))}{2\Phi(\gamma_s^j(\vec{x}))} - \ln \Phi(\gamma_s^j(\vec{x})) \right] \label{eqn_final}
 \end{align} 
A complete description of the MESMO algorithm is given in Algorithm \ref{alg:MESMO}. The blue colored steps correspond to computation of our output space entropy based acquisition function.

\begin{algorithm}[h]
\caption{MESMO Algorithm}
\label{alg:MESMO}
\textbf{Input}: input space $\mathfrak{X}$; $K$ blackbox objective functions $f_1(x), f_2(x),\cdots,f_K(x)$; and maximum no. of iterations $T_{max}$
\begin{algorithmic}[1] 
\STATE Initialize Gaussian process models $\mathcal{GP}_1,\cdots, \mathcal{GP}_K$ by evaluating at $N_0$ initial points
\FOR{each iteration $t$ = $N_0+1$ to $T_{max}$}
\STATE Select $\vec{x}_{t} \leftarrow \arg max_{\vec{x}\in \mathfrak{X}} \hspace{2 mm} \alpha_t(\vec{x}) $, where $\alpha_t(.)$ is computed as:
 \STATE \quad for each sample $s \in {1,\cdots,S}$: 
 \STATE \quad \quad Sample $\Tilde{f}_j \sim \mathcal{GP}_j, \quad \forall{j \in \{1,\cdots, K\}} $
 \STATE \quad \quad $\mathcal{Y}_s^* \leftarrow$ Pareto front of {\em cheap} multi-objective optimization over $(\Tilde{f}_1, \cdots, \Tilde{f}_K)$
 \STATE \quad Compute $\alpha_t$(.) based on the $S$ samples of $\mathcal{Y}_s^*$ as given in equation (\ref{eqn_final})
\STATE Evaluate $\vec{x}_{t}$: $\vec{y}_{t} \leftarrow (f_1(\vec{x}_{t}),\cdots,f_K(\vec{x}_{t}))$ 
\STATE Aggregate data: $\mathcal{D} \leftarrow \mathcal{D} \cup \{(\vec{x}_{t}, \vec{y}_{t})\}$ 
\STATE Update models $\mathcal{GP}_1, \mathcal{GP}_2,\cdots, \mathcal{GP}_K$ 
\STATE $t \leftarrow t+1$
\ENDFOR
\STATE \textbf{return} Pareto front of $f_1(x), f_2(x),\cdots,f_K(x)$ based on $\mathcal{D}$
\end{algorithmic}
\end{algorithm}
\begin{figure}[t]
    \centering
    \includegraphics[width=0.75\columnwidth]{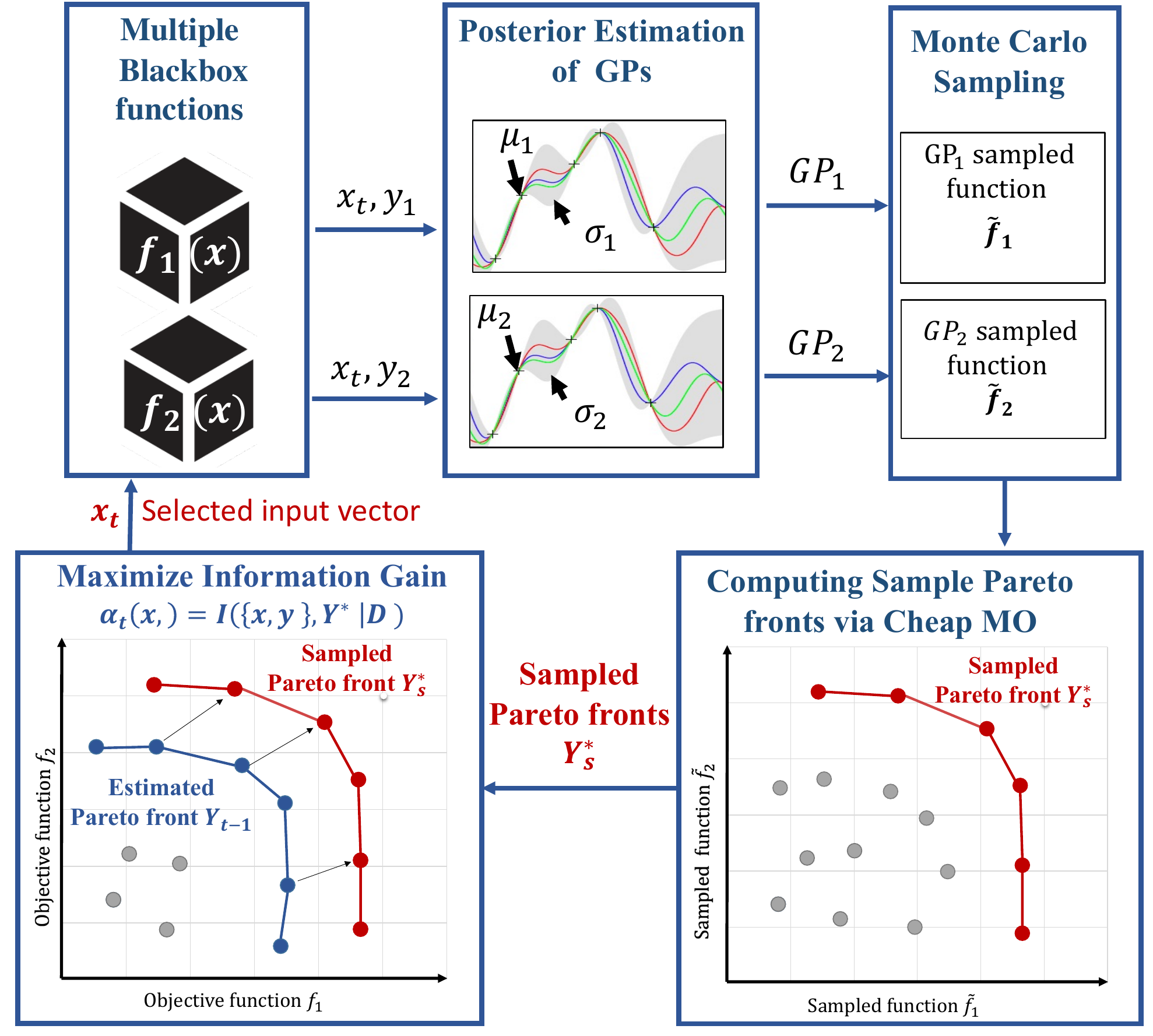}
    \caption{Overview of the MESMO algorithm for two objective functions ($K$=2). We build statistical models $\mathcal{GP}_1$, $\mathcal{GP}_2$ for the two objective functions $f_1(x) $ and $f_2(x)$ respectively. First, we sample functions from the statistical models. We compute sample pareto fronts by solving a cheap MO problem over the sampled functions. Second, we select the best candidate input $x_t$ that maximizes the information gain. Finally, we evaluate the functions for $x_t$ to get $(y_1, y_2)$ and update the statistical models using the new training example.}
    \label{fig:algo_mesmo}
\end{figure} 

%% file: MESMOC_setup_Approach.tex
\section{MESMOC Algorithm for MOO Problem with Constraints}\label{section-MESMOC}
In this section, we address the MOO problem with constraints, where the goal is to optimize multiple real-valued objective functions while satisfying several black-box constraints over continuous space. To solve this problem, we propose an algorithm referred to as {\em {\bf M}ax-value {\bf E}ntropy {\bf S}earch for {\bf M}ulti-objective {\bf O}ptimization with {\bf C}onstraints} (MESMOC). In what follows, we explain the technical details and acquisition function derivation.

\vspace{1.0ex}

\noindent {\bf Problem Setup (MOO Problem with Constraints).}
This is a generalization of the basic MOO problem, where we need to satisfy some black-box constraints. Our goal is to maximize real-valued objective functions $f_1(\vec{x}), f_2(\vec{x}),\cdots,f_K(\vec{x})$, with  $K \geq 2$,  while satisfying $L$ black-box constraints of the form $C_1(x)\geq 0, C_2(x)\geq 0,\cdots,C_L(x)\geq 0$ over continuous space  $\mathfrak{X} \subseteq \Re^d$. Each evaluation of an input $\vec{x}\in \mathfrak{X}$ produces a vector of objective values and constraint values $\vec{y}$ = $(y_{f_1}, y_{f_2},\cdots,y_{f_K},y_{c_1} \cdots y_{c_L})$ where $y_{f_j} = f_j(x)$ for all $j \in \{1,2, \cdots, K\}$ and $y_{c_i} = C_i(x)$ for all $i \in \{1,2, \cdots, L\}$.We say that a valid input $\vec{x}$ (satisfies all constraints) {\em Pareto-dominates} another input $\vec{x'}$ if $f_j(\vec{x}) \geq f_j(\vec{x'}) \hspace{1mm} \forall{j}$ and there exists some $j \in \{1, 2, \cdots,K\}$ such that $f_j(\vec{x}) > f_j(\vec{x'})$. The goal of multi-objective BO with constraints is to approximate the Pareto set over valid inputs $\mathcal{X}^*$ while minimizing the number of expensive function evaluations. For example, in electric aviation power system design applications,
we need to find the designs that trade-off total energy and the mass while satisfying
specific thresholds for motor temperature and voltage of cells. Table~\ref{table:notations} contains all the mathematical notations used in this section.

\vspace{1.0ex}

\noindent {\bf Surrogate Models.}  Similar to section \ref{Section-MESMO}, we model the objective functions and black-box constraints by independent GP models $\mathcal{GP}_{f_1},\mathcal{GP}_{f_2},\cdots,\mathcal{GP}_{f_K}$ and $\mathcal{GP}_{c_1},\mathcal{GP}_{c_2},\cdots,\mathcal{GP}_{f_K}$ with zero mean and i.i.d. observation noise. Let $\mathcal{D} = \{(\vec{x}_i, \vec{y}_i)\}_{i=1}^{t{-1}}$ be the training data from past $t{-1}$ function evaluations, where  
$\vec{x}_i \in \mathfrak{X}$ is an input and $\vec{y}_i = \{y_{f_1}^i,\cdots,y_{f_K}^i,y_{c_1}^i,\cdots y_{c_L}^i\}$ is the output vector resulting from evaluating the objective functions and constraints at $\vec{x}_i$. We learn surrogate models from $\mathcal{D}$.

\subsection{MESMOC Algorithm}
\noindent{\bf Output Space Entropy Based Acquisition Function.}
To overcome the challenges of computing input space entropy based acquisition function,  MESMO \cite{belakaria2019max} proposed to maximize the information gain about the optimal {\bf Pareto front}. However, MESMO did not address the challenge of constrained Pareto front. We propose an extension of MESMO's acquisition function to maximize the information gain between the next candidate input for evaluation $\vec{x}$ and constrained Pareto front $\mathcal{Y}^*$ given as:
\begin{align}
        \alpha(\vec{x}) &= I(\{\vec{x}, \vec{y}\}, \mathcal{Y}^* \mid D) = H(\vec{y} \mid D, \vec{x}) - \mathbb{E}_{\mathcal{Y}^*} [H(\vec{y} \mid D,   \vec{x}, \mathcal{Y}^*)] \label{eqn_symmetric_mesmo1}
\end{align}
In this case, the output vector $\vec{y}$ is $K+L$ dimensional: $\vec{y}$ = $(y_{f_1}, y_{f_2},\cdots,y_{f_K},y_{c_1} \cdots y_{c_L})$ where $y_{f_j}$ = $f_j(x)$ for all $j \in \{1,2, \cdots, K\}$ and $y_{c_i}$ = $C_i(x)$ for all $i \in \{1,2, \cdots, L\}$. 
Consequently, the first term in the  r.h.s of equation (\ref{eqn_symmetric_mesmo1}), entropy of a factorizable $(K+L)$-dimensional Gaussian distribution $P(\vec{y}\mid D, \vec{x}$, can be computed in closed form as shown below:
\begin{align}
    H(\vec{y} \mid D, \vec{x}) = \frac{(K+C)(1+\ln(2\pi))}{2} +  \sum_{j = 1}^K  \ln (\sigma_{f_j}(\vec{x}))+  \sum_{i = 1}^L  \ln (\sigma_{c_i}(\vec{x})) \label{eqn_unconditioned_entropy1}
\end{align}
where $\sigma_{f_j}^2(\vec{x})$ and  $\sigma_{c_i}^2(\vec{x})$ are the predictive variances of $j^{th}$ function and $i^{th}$ constraint GPs respectively at input $\vec{x}$.

The l.h.s of equation (\ref{eqn_symmetric_mesmo1}) can be decomposed in a similar way to equation (\ref{eqn_summation}). There are two key algorithmic steps to compute this part of the equation: 1) The first is how to compute Pareto front samples $\mathcal{Y}^*_s$?; and 2) The second is how to compute the entropy with respect to a given Pareto front sample $\mathcal{Y}^*_s$? We provide solutions for these two questions below.

\vspace{1.0ex}

\hspace{2.0ex} {\bf 1) Computing Pareto Front Samples via Cheap Multi-Objective Optimization.} To compute a Pareto front sample $\mathcal{Y}^*_s$, we first sample functions and constraints from the posterior GP models via random Fourier features \cite{PES,random_fourier_features} and then solve a cheap constrained multi-objective optimization over the $K$ sampled functions and $L$ sampled constraints.

\hspace{3.5ex}{\em Cheap MO solver.} We sample $\Tilde{f}_i$ from GP model $\mathcal{GP}_{f_j}$ for each of the $K$ functions and $\Tilde{C}_i$ from GP model $\mathcal{GP}_{c_i}$ for each of the $L$ constraints. A {\em cheap} constrained multi-objective optimization problem over the $K$ sampled functions  $\Tilde{f}_1,\Tilde{f}_2,\cdots,\Tilde{f}_k$ and the $L$ sampled constraints  $\Tilde{C}_1,\Tilde{C}_2,\cdots,\Tilde{C}_L$ is solved to compute the sample Pareto front $\mathcal{Y}^*_s$. We employ the popular constrained NSGA-II algorithm \cite{deb2002nsga,deb2002fast} to solve the constrained MO problem with cheap sampled objective functions and constrained.

\vspace{1.0ex}

\hspace{2.0ex}{\bf 2) Entropy Computation with a Sample Pareto Front.}
Let $\mathcal{Y}^*_s = \{\vec{v}^1, \cdots, \vec{v}^l \}$ be the sample Pareto front,  where $l$ is the size of the Pareto front and each $\vec{v}^i$ is a $(K+L)$-vector evaluated at the $K$ sampled functions and $L$ sampled constraints $\vec{v}^i = \{v^i_{f_1},\cdots,v^i_{f_K},v^i_{c_1},\cdots,v^i_{c_L}\}$. The following inequality holds for each component $y_j$ of the $(K+L)$-vector $\vec{y} = \{y_{f_1},\cdots,y_{f_K},y_{c_1},\cdots y_{c_L}\}$ in the entropy term $H(\vec{y} \mid D,   \vec{x}, \mathcal{Y}^*_s)$:
\begin{align}
 y_j &\leq \max \{v^1_j, \cdots v^l_j \} \quad \forall j \in \{f_1,\cdots,f_K,c_1,\cdots,c_L\} \label{inequality1}
\end{align}

The inequality essentially says that the $j^{th}$ component of $\vec{y}$ (i.e., $y_j$) is upper-bounded by a value obtained by taking the maximum of $j^{th}$ components of all $l$ $(K+L)$-vectors in the Pareto front $\mathcal{Y}^*_s$. This inequality had been proven by a contradiction for MESMO \cite{belakaria2019max} for $j \in \{f_1,\cdots,f_K\}$. We assume the same  for $j \in \{c_1,\cdots,c_L\}$.

By combining the inequality (\ref{inequality1}) and the fact that each function is modeled as an independent GP, we can approximate each component $y_j$ as a truncated Gaussian distribution since the distribution of $y_j$ needs to satisfy $ y_j \leq \max \{v^1_j, \cdots v^l_j \}$. Let $y_s^{c_i*} = \max \{v^1_{c_i}, \cdots v^l_{c_i} \}$ and $y_s^{f_j*} = \max \{v^1_{f_j}, \cdots v^l_{f_j} \}$. Furthermore, a common property of entropy measure allows us to decompose the entropy of a set of independent variables into a sum over entropies of individual variables \cite{information_theory}:
\begin{align}
H(\vec{y} \mid D,   \vec{x}, \mathcal{Y}^*_s) = \sum_{j=1}^K H(y_{f_j}|D, \vec{x}, y_s^{f_j*}) +\sum_{i=1}^C H(y_{c_i}|D, \vec{x}, y_s^{c_i*})  \label{eqn_sep_ineq1}
\end{align}
The r.h.s is a summation over entropies of  $(K+L)$-variables $\vec{y} = \{y_{f_1},\cdots,y_{f_K},y_{c_1},\cdots y_{c_L}\}$.
The differential entropy for each $y_j$ is the entropy of a truncated Gaussian distribution \cite{entropy_handbook} and given by the following equations:
\begin{align}
    H(y_{f_j}|D, \vec{x}, y_s^{f_j*}) &\simeq  \left[\frac{(1 + \ln(2\pi))}{2}+  \ln(\sigma_{f_j}(\vec{x})) +  \ln \Phi(\gamma_s^{f_j}(\vec{x})) - \frac{\gamma_s^{f_j}(\vec{x}) \phi(\gamma_s^{f_j}(\vec{x}))}{2\Phi(\gamma_s^{f_j}(\vec{x}))}\right] \label{eqn_entropy_fj}
  \end{align}
\begin{align}
   H(y_{c_i}|D, \vec{x}, y_s^{c_i*})&\simeq  \left[\frac{(1 + \ln(2\pi))}{2}+  \ln(\sigma_{c_i}(\vec{x})) +  \ln \Phi(\gamma_s^{c_i}(\vec{x})) - \frac{\gamma_s^{c_i}(\vec{x}) \phi(\gamma_s^{c_i}(\vec{x}))}{2\Phi(\gamma_s^{c_i}(\vec{x}))}\right]    
    \label{eqn_entropy_cj}
\end{align} 
Consequently we have: 
\begin{align}
    H(\vec{y} \mid D,   \vec{x}, \mathcal{Y}^*_s) &\simeq \sum_{j=1}^K \left[\frac{(1 + \ln(2\pi))}{2}+  \ln(\sigma_{f_j}(\vec{x})) +  \ln \Phi(\gamma_s^{f_j}(\vec{x})) - \frac{\gamma_s^{f_j}(\vec{x}) \phi(\gamma_s^{f_j}(\vec{x}))}{2\Phi(\gamma_s^{f_j}(\vec{x}))}\right] \nonumber\\
&+ \sum_{i=1}^L \left[\frac{(1 + \ln(2\pi))}{2}+  \ln(\sigma_{c_i}(\vec{x})) +  \ln \Phi(\gamma_s^{c_i}(\vec{x})) - \frac{\gamma_s^{c_i}(\vec{x}) \phi(\gamma_s^{c_i}(\vec{x}))}{2\Phi(\gamma_s^{c_i}(\vec{x}))}\right]    
    \label{eqn_entropy_closed1}
\end{align}

where $\gamma_s^{c_i}(x) = \frac{y_s^{c_i*} - \mu_{c_i}(\vec{x})}{\sigma_{c_i}(\vec{x})}$, $\gamma_s^{f_j}(x) = \frac{y_s^{f_j*} - \mu_{f_j}(\vec{x})}{\sigma_{f_j}(\vec{x})}$, 
$\phi$ and $\Phi$ are the p.d.f and c.d.f of a standard normal distribution respectively. By combining equations (\ref{eqn_unconditioned_entropy1}) and (\ref{eqn_entropy_closed1}) with equation (\ref{eqn_symmetric_mesmo1}), we get the final form of our acquisition function as shown below:
\begin{align}
\alpha(\vec{x}) \simeq \frac{1}{S} \sum_{s=1}^S\left[ \sum_{j=1}^K  \frac{\gamma_s^{f_j}(\vec{x}) \phi(\gamma_s^{f_j}(\vec{x}))}{2\Phi(\gamma_s^{f_j}(\vec{x}))} - \ln \Phi(\gamma_s^{f_j}(\vec{x}))  + \sum_{i=1}^L  \frac{\gamma_s^{c_i}(\vec{x}) \phi(\gamma_s^{c_i}(\vec{x}))}{2\Phi(\gamma_s^{c_i}(\vec{x}))} - \ln \Phi(\gamma_s^{c_i}(\vec{x})) \right] \label{eqn_final1}
 \end{align} 
A complete description of the MESMOC algorithm is given in Algorithm \ref{alg:MESMOC}.

\begin{algorithm}[h]
\caption{MESMOC Algorithm}
\label{alg:MESMOC}
\textbf{Input}: input space $\mathfrak{X}$; $K$ blackbox functions $f_1(x), f_2(x),\cdots,f_K(x)$; $L$ blackbox constraints $C_1(x), C_2(x),\cdots,C_L(x)$; and  maximum no. of iterations $T_{max}$
\begin{algorithmic}[1] 
\STATE Initialize Gaussian process models $\mathcal{GP}_{f_1}, \mathcal{GP}_{f_2},\cdots, \mathcal{GP}_{f_K}$ and $\mathcal{GP}_{c_1}, \mathcal{GP}_{c_2},\cdots, \mathcal{GP}_{c_L}$ by evaluating at $N_0$ initial points
\FOR{each iteration $t$ = $N_0+1$ to $T_{max}$}
\STATE Select $\vec{x}_{t} \leftarrow \arg max_{\vec{x}\in \mathfrak{X}} \hspace{2 mm} \alpha_t(\vec{x}) $
 \\ \qquad \qquad \textbf{s.t} $(\mu_{c_1}\geq 0, \cdots,\mu_{c_L}\geq 0 )$
 \STATE $\alpha_t(.)$ is computed as:
 \STATE \quad for each sample $s \in {1,\cdots,S}$: 
 \STATE \quad \quad Sample $\Tilde{f}_j \sim \mathcal{GP}_{f_j}, \quad \forall{j \in \{1,\cdots, K\}} $
  \STATE \quad \quad Sample $\Tilde{C}_i \sim \mathcal{GP}_{c_i}, \quad \forall{i \in \{1,\cdots, L\}} $
 \STATE \quad \quad // Solve {\em cheap} MOO over $(\Tilde{f}_1, \cdots, \Tilde{f}_K)$ constrained by $(\Tilde{C}_1, \cdots, \Tilde{C}_L)$
 \STATE \quad \quad $\mathcal{Y}_s^* \leftarrow \arg max_{x \in \mathcal{X}} (\Tilde{f}_1, \cdots, \Tilde{f}_K)$ \\ \qquad \qquad \textbf{s.t} $(\Tilde{C}_1\geq 0, \cdots, \Tilde{C}_L\geq 0)$ 
 \STATE  \quad Compute $\alpha_t$(.) based on the $S$ samples of $\mathcal{Y}_s^*$ as given in equation (\ref{eqn_final1})
\STATE Evaluate $\vec{x}_{t}$; $\vec{y}_{t} \leftarrow (f_1(\vec{x}_{t}),\cdots,f_K(\vec{x}_{t}),C_1(\vec{x}_{t}),\cdots,C_L(\vec{x}_{t}))$ 
\STATE Aggregate data: $\mathcal{D} \leftarrow \mathcal{D} \cup \{(\vec{x}_{t}, \vec{y}_{t})\}$ 
\STATE Update models $\mathcal{GP}_{f_1}, \mathcal{GP}_{f_2},\cdots, \mathcal{GP}_{f_K}$ and $\mathcal{GP}_{c_1}, \mathcal{GP}_{c_2},\cdots, \mathcal{GP}_{c_L}$
\STATE $t \leftarrow t+1$
\ENDFOR
\STATE \textbf{return} Pareto front of $f_1(x), f_2(x),\cdots,f_K(x)$ based on $\mathcal{D}$
\end{algorithmic}
\end{algorithm}

%% file: MF_OESMO_setup_Approach.tex
\section{MF-OSEMO Algorithm for Discrete Multi-Fidelity MOO Problem}

In this section, we address the multi-fidelity version of MOO problem, where we have access to multiple fidelities for each function that vary in the amount of resources consumed and the accuracy of evaluation. To solve this problem, we propose an algorithm referred to as {\em {\bf M}ulti-{\bf F}idelity {\bf O}utput {\bf S}pace {\bf E}ntropy Search for {\bf M}ulti-objective {\bf O}ptimization } (MF-OSEMO). We first describe the complete details related to the multi-fidelity MOO problem. Subsequently, we explain our proposed MF-OSEMO algorithm with two mathematically different approximations of the output space entropy based acquisition function.

\vspace{1.0ex}

\noindent {\bf Problem Setup (Discrete Multi-Fidelity MOO Problem)}.
This is a general version of the MOO problem, where we have access to $M_j$ fidelities for each function $f_j$ that vary in the amount of resources consumed and the accuracy of evaluation. The evaluation of an input $\vec{x}\in \mathfrak{X}$ with fidelity vector $\vec{m} = [m_1, m_2, \cdots, m_K]$ produces an evaluation vector of $K$ values denoted by $\vec{y}^{\vec{m}} \equiv [y_1^{(m_1)}, \cdots, y_K^{(m_K)}]$, where $y_j^{(m_j)} = f_j^{(m_j)}(x)$ for all $j \in \{1,2, \cdots, K\}$. Let $\lambda_j^{(m_j)}$ be the cost of evaluating $i^{th}$ function $f_j$ at $m_j \in [M_j]$ fidelity, where $m_j$=$M_j$ corresponds to the highest fidelity for $f_j$. Our goal is to approximate the optimal Pareto set $\mathcal{X}^*$ over the highest fidelities functions while minimizing the overall cost of function evaluations (experiments). For example, in power system design optimization, we need to find designs that trade-off cost, size, efficiency, and thermal tolerance using multi-fidelity simulators for design evaluations. Table~\ref{table:notations2} contains all the mathematical notations used in this section (MF-OSEMO).

\vspace{1.0ex}

\noindent{\bf Cost of Function Evaluations.} The total normalized evaluation cost is \\ $\lambda^{(\vec{m})} \equiv \sum_{j=1}^{K} \left({\lambda_j^{(m_j)}}/{\lambda_j^{(M_j)}}\right)$. We normalize the total cost since the cost units can be different for different objectives (e.g. cost unit for $f_1$ is computation time while cost unit for $f_2$ could be memory space size). If the cost is known, it can be directly injected in the latter expression. However, in some real world settings, the cost of a function evaluation can be only known after the function evaluation. For example, in hyper-parameter tuning of a neural network, the cost of the experiment is defined by the training and inference time. However, we cannot know the exact needed time until after the experiment is finalised. In this case, the cost can be modeled by an independent Gaussian process. The predictive mean can be used during the optimization. 
Our goal is to approximate $\mathcal{X}^*$ by minimizing the overall cost of function evaluations. 

\begin{table*}[t]
    \centering
    \resizebox{1\linewidth}{!}{
    \begin{tabular}{|c|c|}
    \hline
    {\bf Notation} & {\bf Definition} \\
    \hline \hline
      $f_1^{(m_1)}, f_2^{(m_1)}, \cdots, f_K^{(m_K)}$  & functions the $m_j$ fidelity of the true objective functions \\
      \hline
    $\Tilde{f}_j^{(m_j)}$ & function sampled from $j$th Gaussian process model at $m_j$th fidelity\\
\hline
       $M_1, M_2, \cdots, M_K$ & no. of fidelities for each function \\
       \hline
    $\vec{m} = [m_1, m_2, \cdots, m_K]$& fidelity vector where each fidelity $m_j \in [M_j]$ \\
       \hline
        $y_j^{m_j}$ &$j$th function $f_j$ evaluated at $m_j$th fidelity  where $m_j \in [M_j]$\\
       \hline
      $\vec{y}^{\vec{m}}$   & output vector equivalent to $ [y_1^{(m_1)}, \cdots, y_K^{(m_K)}]$\\  
\hline 
$\lambda_j^{({m_j})}$ & cost of evaluating $j$th function $f_j$ at $m_j$th fidelity\\
\hline
$\lambda^{(\vec{m})}$ & total normalized cost $\lambda^{(\vec{m})} \equiv \sum_{j=1}^{K} \left( {\lambda_j^{(m_j)}}/{\lambda_j^{(M_j)}}\right)$  \\
\hline
$\mathcal{Y}^*$ & true pareto front of the objective functions $[f_1, f_2, \cdots, f_K]$ (the highest fidelities )\\      \hline 
$\mathcal{Y}_s^*$ &  Pareto front of the sampled highest fidelities $[\Tilde{f}_1, \Tilde{f}_2, \cdots, \Tilde{f}_K]$\\ 
\hline
    \end{tabular}}
    \caption{Table describing additional mathematical notations used in this section (MF-OSEMO).}
    \label{table:notations2}
\end{table*}

\vspace{1.0ex}

\noindent\textbf{Multi-Fidelity Gaussian Process Model.} \label{surrogatesection2} Let $D = \{(\vec{x}_i, \vec{y}_i^{(\vec{m})})\}_{i=1}^{t-1}$ be the training data from past $t{-1}$ function evaluations, where  
$\vec{x}_i \in \mathfrak{X}$ is an input and $\vec{y}^{(\vec{m})}_i = [y_1^{(m_1)},\cdots,y_K^{(m_K)}]$ is the output vector resulting from evaluating functions $f_1^{(m_1)}, f_2^{(m_2)},\cdots,f_K^{(m_k)}$ at $\vec{x}_i$. Gaussian processes (GPs) are known to be effective surrogate models in prior work on single and multi-objective BO \cite{gp-ucb,PESMO}. 
We learn $K$ surrogate models $\mathcal{GP}_1,\mathcal{GP}_2,\cdots,\mathcal{GP}_K$ from $\mathcal{D}$, where each $\mathcal{GP}_j$ corresponds to the $j$th function $f_j$. In our setting, each function has multiple fidelities. So one ideal property desired for the surrogate model of a single function is to take into account all the fidelities in a single model. Multi-fidelity GPs (MF-GP) are capable of modeling functions with multiple fidelities in a single model. Hence, each of our surrogate model $\mathcal{GP}_j$ is a multi-fidelity GP.

Specifically, we use the MF-GP model as proposed in \citeA{kennedy2000predicting,takeno2019multi}. We describe the complete details of the MF-GP model below for the sake of completeness. One key thing to note about MF-GP model is that the kernel function ($k((\vec{x_i},m_i),(\vec{x_j},m_j))$) is dependent on both the input and the fidelity. 
For a given input $\vec{x}$, the MF-GP model returns a {\em vector} (one for each fidelity) of predictive mean, a {\em vector} of predictive variance, and a matrix of predictive covariance.
The MF-GP model has two advantages. The first is that all fidelities are integrated into one single GP.  The second is that difference among fidelities are adaptively estimated without any additional feature representation for fidelities. It should be noted that we employ an independent multi-fidelity GP for each function.

\vspace{2.0ex}

\noindent We describe full details of a MF-GP model for one objective function $f_j$ (without loss of generality) below:

Let $y_j^{(1)}(\vec{x}), \ldots, y_j^{(M_j)}(\vec{x})$ represent the values obtained by evaluating the function $f_j$  at its $1$st, $2$nd, $\ldots, M_j$th  fidelity respectively. In a MF-GP model, each fidelity is represented by a Gaussian process and the observation is modeled as
\begin{align*}
 y_j^{(m_j)}(\vec{x}) = f_j^{(m_j)}(\vec{x}) + \epsilon, \quad \epsilon \sim \mathcal{N}(0, \sigma_{\rm noise}^2).
\end{align*}
Let $f_j^{(1)} \sim GP(0,k_1(\vec{x},\vec{x}'))$
be a Gaussian process for the $1$st fidelity i.e. $m_j = 1$, where $k_1: \mathcal{R}^{d} \times\mathcal{R}^{d} \rightarrow \mathcal{R}$ is a suitable kernel.
The output for successively fidelities $m_j = 2, \ldots, M_j$ is recursively defined as
\begin{align}
 f_j^{(m_j)}(\vec{x}) & = f_j^{(m_j-1)}(\vec{x}) + f_{j_e}^{(m_j-1)}(\vec{x}), 
 \label{eq:fidelity-diff}
\end{align}
where, $f_{j_e}^{(m_j-1)} \sim GP(0, k_e(\vec{x},\vec{x}^\prime))$
with $k_e: \mathcal{R}^d \times \mathcal{R}^d \rightarrow \mathcal{R}$.
It is assumed that $f_{j_e}^{(m_j-1)}$ is conditionally independent from all fidelities lower than $m_j$. As a result, {\em the kernel for a pair of points evaluated at the same fidelity} becomes:
\begin{align}
 k_{m_j}(\vec{x},\vec{x}^\prime) \equiv k_1(\vec{x},\vec{x}^\prime) + (m_j-1) k_e(\vec{x},\vec{x}^\prime)
 \label{eq:mfgpr-kernel}
\end{align}
and as a result, the output for $m_j$th fidelity is also modeled as a Gaussian process:
\begin{align*}
 f_j^{(m_j)} \sim GP(0, k_{m_j}(\vec{x},\vec{x}^\prime)).
\end{align*}

\noindent{\em The kernel function for a pair of inputs evaluated at different fidelities $m_j$ and $m_j'$ is}:
\begin{align*}
 k((\vec{x},m_j),(\vec{x}',m_j')) 
 & =
 \mathrm{cov}\left(
 f_j^{(m_j)}(\vec{x}),
 f_j^{(m_j')}(\vec{x}')
 \right)=
 k_{m_j}(\vec{x},\vec{x}')
\end{align*}
where $m_j \leq m_j'$ and $\mathrm{cov}$ represents covariance.
Using a kernel matrix $\*K \in \mathcal{R}^{n \times n}$ in which the $p,q$ element is defined by 
$k((\vec{x},m_j^p),(\vec{x}',m_j^q))$, 
all fidelities $f_j^{(1)}, \ldots, f_j^{(M_j)}$ can be integrated into one common Gaussian process by which predictive mean and variance are obtained as 
\begin{align}
 \mu^{(m_j)}(\vec{x}) 
 & = 
 \*K + \sigma_{\rm noise}^2 \*I^{-1}
 \vec{Y}, 
 \label{eq:MF-GPR-mean}
 \\
 \sigma^{2^{(m_j)}}(\vec{x})  
 & = 
 k((\vec{x},m_j),(\vec{x},m_j))  
 - \*k^{(m_j)}_n(\vec{x})^\top 
 \*K + \sigma_{\rm noise}^2 \*I^{-1}
 \*k^{(m_j)}_n(\vec{x}),
 \label{eq:MF-GPR-var}
\end{align}
where $k^{(m_j)}_n(\vec{x}) \equiv (k((\vec{x},m_j),(\vec{x}_1,{m_j}_1))
 , \ldots, k((\vec{x},m_j),(\vec{x}_n,{m_j}_n)))^\top$ and\\  $\vec{Y} = (y_1^{(m_{j_1})}(\vec{x}_1), \ldots, y_n^{(m_{j_n})}(\vec{x}_n))^\top$. We also define  $\sigma^{2^{(m_jm_j^\prime)}}(\vec{x})$ as the predictive covariance between $(\vec{x},m_j)$ and $(\vec{x},m_j^\prime)$, i.e., covariance for identical $\vec{x}$ at different fidelities:

\begin{align}
  \sigma^{2(m_jm_j^\prime)} & (\vec{x}) 
   = 
  k((\vec{x},m_j),(\vec{x},m_j^\prime))  -  \*k^{(m_j)}_n(\vec{x})^\top 
 \*K + \sigma_{\rm noise}^2 \*I^{-1}
 \*k^{(m_j^\prime)}_n(\vec{x}).
 \label{eq:MF-GPR-cov}
\end{align}

\subsection{MF-OSEMO Algorithm with Two Approximations}

We describe our proposed acquisition function for the multi-fidelity MOO problem setting. We leverage the information-theoretic principle of output space information gain to develop an efficient and robust acquisition function. This method is applicable for the general case, where at each iteration, the objective functions can be evaluated at different fidelities. 

\begin{figure}[t]
    \centering
    \includegraphics[width=0.9\columnwidth]{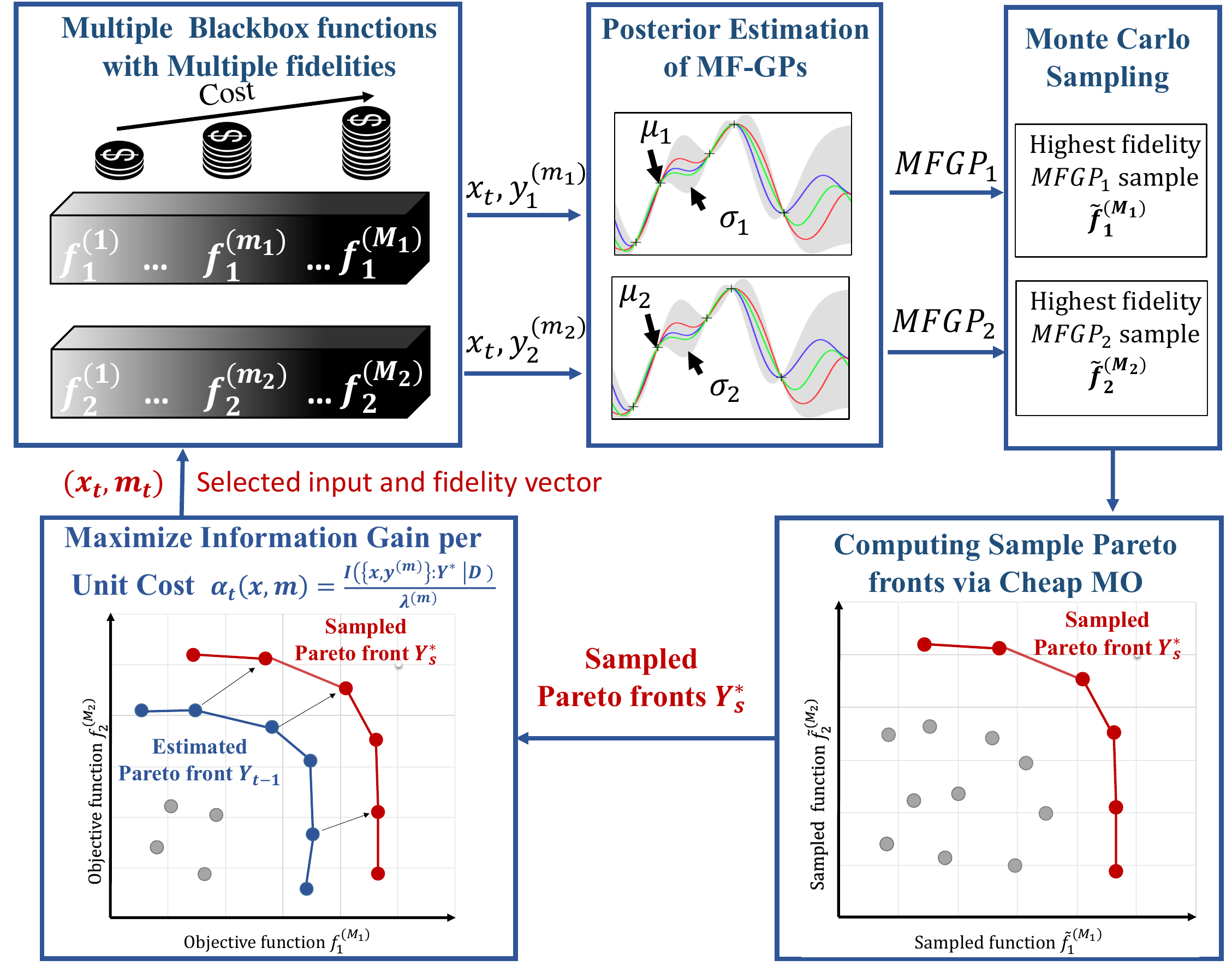}
    \caption{Overview of the MF-OSEMO algorithm for two objective functions ($K$=2). We build multi-fidelity statistical models $\mathcal{MFGP}_1$, $\mathcal{MFGP}_2$ for the two objective functions $f_1(x) $ and $f_2(x)$ with $M_1$ and $M_2$ fildelities respectively. First, we sample highest fidelity functions from the statistical models. We compute sample pareto fronts by solving a cheap MO problem over the sampled functions. Second, we select the best candidate input $x_t$ and fidelity vector $m_t=(m_1,m_2)$ that maximizes the information gain per unit cost . Finally, we evaluate the functions for $x_t$ at fidelities $m_t$ to get $(y_1^{(m_1)}, y_2^{(m_2)})$ and update the statistical models using the new training example.}
    \label{fig:algo}
\end{figure}

The key idea behind the proposed acquisition function is to find the pair $\{\vec{x}, \vec{m}\}$ that maximizes the information gain about the {\bf Pareto front of the highest fidelities (denoted by $\mathcal{Y}^*$)} per unit cost, where $\{\vec{x}, \vec{m}\}$ represents a candidate input $\vec{x}$ evaluated at a vector of fidelities $\vec{m} = [m_1, m_2, \cdots, m_K]$. This idea  can be expressed mathematically 
as given below:
\begin{align}
        \alpha(\vec{x},\vec{m}) &= I(\{\vec{x}, \vec{y}^{(\vec{m})}\}, \mathcal{Y}^* \mid D) / \lambda^{(\vec{m})} \label{af:def2}
\end{align}
where $\lambda^{(\vec{m})}$ is the total {\em normalized} cost of evaluating the objective functions at  $\vec{m}$ and $D$ is the data collected so far.
Figure \ref{fig:algo} provides an overview of the  MF-OSEMO algorithm.
The information gain in equation (\ref{af:def2}) is defined as the expected reduction in entropy $H(.)$ of the posterior distribution $P(\mathcal{Y}^* \mid D)$ as a result of evaluating $\vec{x}$ at fidelity vector $\vec{m}$:
\begin{align}
    I(\{\vec{x}, \vec{y}^{(\vec{m})}\}, \mathcal{Y}^{*} \mid D) &= H(\mathcal{Y}^{*} \mid D) - \mathbb{E}_{y^{(\vec{m})}} [H(\mathcal{Y}^{*} \mid D \cup \{\vec{x}, \vec{y}^{(\vec{m})}\})] \label{eqn_ig_def} \\
    &= H(\vec{y}^{(\vec{m})} \mid D, \vec{x}) - \mathbb{E}_{\mathcal{Y}^{*}} [H(\vec{y}^{(\vec{m})} \mid D,   \vec{x}, \mathcal{Y}^{*})]  \label{eqn_symmetric_ig}
\end{align}
equation (\ref{eqn_symmetric_ig}) follows from equation (\ref{eqn_ig_def}) as a result of the symmetric property of information gain. The first term in the  r.h.s of equation (\ref{eqn_symmetric_ig}) is the entropy of a factorizable K-dimensional Gaussian distribution $P(\vec{y}^{(\vec{m})}\mid D, \vec{x}$)) which can be computed in closed form as shown below:
\begin{align}
H(\vec{y}^{(\vec{m})} \mid D, \vec{x}) = \frac{K(1+\ln(2\pi))}{2} +  \sum_{j = 1}^K  \ln (\sigma_j^{(m_j)}(\vec{x})) \label{eqn_unconditioned_entropy2}
\end{align}
where $\sigma_j^{{(m_j)}}(\vec{x})$ is the predictive variance of $j^{th}$ surrogate model $GP_j$ at input $\vec{x}$ and fidelity $m_j$. The second term in the r.h.s of equation (\ref{eqn_symmetric_ig}) is an expectation over the Pareto front of the highest fidelities $\mathcal{Y}^{*}$. We can approximately compute this term via Monte-Carlo sampling as shown below: 

\begin{align}
    \mathbb{E}_{\mathcal{Y}^{*}} [H(\vec{y}^{(\vec{m})} \mid D,   \vec{x}, \mathcal{Y}^{*})] \simeq \frac{1}{S} \sum_{s = 1}^S [H(\vec{y}^{(\vec{m})} \mid D,   \vec{x}, \mathcal{Y}^{*}_s)] \label{eqn_summation2}
\end{align}
where $S$ is the number of samples and $\mathcal{Y}^{*}_s$ denote a sample Pareto front obtained over the highest fidelity functions sample  from $K$ surrogate models. 
The main advantages of our acquisition function are: cost efficiency, computational-efficiency, and robustness to the number of samples. Our experiments demonstrate these advantages over state-of-the-art single fidelity AFs for multi-objective optimization. 

There are two key algorithmic steps to compute equation (\ref{eqn_summation2}). The first is computing Pareto front samples $\mathcal{Y}^{*}_s$; and the second is computing the entropy with respect to a given Pareto front sample $\mathcal{Y}^{*}_s$. We provide solutions for these two steps below.

\vspace{1.0ex}

\hspace{2.0ex} {\bf 1) Computing Pareto Front Samples via Cheap Multi-Objective optimization.} To compute a Pareto front sample $\mathcal{Y}^{*}_s$, we first sample highest fidelity functions from the posterior MF-GP models via random Fourier features \cite{PES,random_fourier_features} and then solve a cheap multi-objective optimization over the $K$ sampled high fidelity functions. It is important to note that we are sampling only the {\bf highest fidelity function} from each MF-GP surrogate model.

\hspace{3.5ex} {\em Sampling functions from  the posterior of MF-GP model.} Similar to prior work \cite{PES,PESMO,MES}, we employ random Fourier features based sampling procedure. We approximate each GP prior of the highest fidelity as $\Tilde{f}^{(M)} = \phi(\vec{x})^T \theta$, where $\theta \sim N(0, \vec{I})$. The key idea behind random Fourier features is to construct each function sample $\Tilde{f}^{(M)}(\vec{x})$ as a finitely parametrized approximation: $\phi(\vec{x})^T \theta$, where $\theta$ is sampled from its corresponding posterior distribution conditioned on the data $D$ obtained from past function evaluations: $\theta | D \sim N(\vec{A^{-1}\Phi^Ty}_n, \sigma^2\vec{A^{-1}})$, where $\vec{A} = \vec{\Phi^T\Phi} + \sigma^2 \vec{I}$ and $\Phi^T = [\phi(\vec{x}_1),\cdots,\phi(\vec{x}_{t-1})]$.

\hspace{3.5ex}{\em Cheap MO solver.} We sample $\Tilde{f}^{(M_i)}_i$ from each surrogate model $\mathcal{MF-GP}_i$ as described above. A {\em cheap} multi-objective optimization problem over the $K$ sampled functions $\Tilde{f}^{(M_1)}_1,\Tilde{f}^{(M_2)}_2,\cdots,\Tilde{f}^{(M_K)}_K$ is solved to compute the sample Pareto front $\mathcal{Y}^{*}_s$. This cheap multi-objective optimization also allows us to capture the interactions between different objectives. We employ the popular NSGA-II algorithm \cite{deb2002nsga} to solve the MO problem with cheap objective functions noting that any other algorithm can be used. 

\vspace{1.0ex}

\hspace{2.0ex}{\bf 2) Entropy Computation with a Sample Pareto Front.}
Let $\mathcal{Y}^{*}_s = \{\vec{v}^1, \cdots, \vec{v}^l \}$ be the sample Pareto front,  where $l$ is the size of the Pareto front and each $\vec{v}^i = \{v_1^i,\cdots,v_K^i\}$ is a $K$-vector evaluated at the $K$ sampled high fidelity functions. The following inequality holds for each component $y^{(m_j)}_j$ of the $K$-vector $\vec{y}^{\vec{(m)}} = \{y^{(m_1)}_1, \cdots, y^{(m_k)}_K\}$ in the entropy term $H(\vec{y}^{\vec{(m)}} \mid D,   \vec{x}, \mathcal{Y}^{*}_s)$:
\begin{align}
 y^{(m_j)}_j &\leq y_{j_s}^{*} \quad \forall j \in \{1,\cdots,K\} \label{inequality2}
\end{align}
where $y_{j_s}^{*} = \max \{v^1_j, \cdots v^l_j \}$. The inequality essentially says that the $j^{th}$ component of $\vec{y}^{\vec{m}}$ (i.e., $y^{m_j}_j$) is upper-bounded by a value obtained by taking the maximum of $j^{th}$ components of all $l$ vectors $\{\vec{v}^1, \cdots, \vec{v}^l \}$ in the Pareto front $\mathcal{Y}^{*}_s$. The proof of \ref{inequality2} can be divided into two cases:

\vspace{0.3ex}

{\bf Case I.} If $y_j$ is evaluated at its highest fidelity (i.e $m_j=M_j$), inequality (\ref{inequality2}) can be proven by a contradiction argument. Suppose there exists some component $y^{(M_j)}_j$ of $\vec{y}^{(\vec{M})}$ such that $ y^{(M_j)}_j > y_{j_s}^{*}$. However, by definition, $\vec{y}^{(\vec{M})}$ is a non-dominated point because no point dominates it in the $j$th dimension. This results in $\vec{y}^{(\vec{M})} \in \mathcal{Y}^*_s$ which is a contradiction. Therefore, our hypothesis that $ y^{(M_j)}_j> y_{j_s}^{*}$ is incorrect and inequality (\ref{inequality2}) holds.

\vspace{0.3ex}

{\bf Case II.} If $y_j$ is evaluated at one of its lower fidelities (i.e, $m_j \neq M_j$), the proof follows from the assumption that the value of lower fidelity of a objective is usually smaller than the corresponding higher fidelity, i.e., $y^{(m_j)}_j \leq y^{(M_j)}_j \leq y_{j_s}^{*}$. This is especially true for most real-world experiments. For example, in optimizing a neural network's accuracy with respect to its hyperparameters, a commonly employed fidelity is the number of data samples used for training. It is reasonable to believe that the accuracy is always higher for the higher fidelity (more data samples to train on) when compared to a lower fidelity (less data samples). By combining the inequality (\ref{inequality2}) and the fact that each function is modeled as an independent MF-GP, a common property of entropy measure allows us to decompose the entropy of a set of independent variables into a sum over entropies of individual variables \cite{information_theory}:
\begin{align}
H(\vec{y}^{\vec{(m)}} \mid D,   \vec{x}, \mathcal{Y}^{*}_s) = \sum_{j=1}^K H(y^{(m_j)}_j|D, \vec{x},y_{j_s}^{*}) \label{eqn_sep_ineq2}
\end{align}
The computation of equation (\ref{eqn_sep_ineq2}) requires the computation of the entropy of $p(y^{(m_j)}_j|D, \vec{x},y_{j_s}^{*})$. This is a conditional distribution that depends on the value of $m_j$ and can be expressed as $H(y^{(m_j)}_j|D, \vec{x},y^{(m_j)}_j\leq y_{j_s}^{*})$. This entropy is dealt with in two cases.

\textbf{First, for $\mathbf{m_j=M_j}$,} the density function of this probability is approximated by truncated Gaussian distribution and its entropy can be expressed as \cite{entropy_handbook}:
\begin{align}
H(y^{(M_j)}_j|D, \vec{x},y^{(M_j)}_j \leq y_{j_s}^{*})\simeq &\frac{(1 + \ln(2\pi))}{2}  +\ln(\sigma^{(M_j)}_j(\vec{x}))  + \ln \Phi(\gamma_s^{(M_j)}(\vec{x})) \nonumber\\
& -\frac{\gamma_s^{(M_j)}(\vec{x}) \phi(\gamma_s^{(M_j)}(\vec{x}))}{2\Phi(\gamma_s^{(M_j)}(\vec{x}))}  \label{Conditional_high_fidel}
\end{align}
where $\gamma_s^{(M_j)}(\vec{x}) = \frac{y_{j_s}^{*} - \mu_j^{(M_j)}(\vec{x})}{\sigma_j^{(M_j)}(\vec{x})}$, and $\phi$ and $\Phi$ are the p.d.f and c.d.f of a standard normal distribution respectively.

\textbf{Second, for $\mathbf{m_j \neq M_j}$}, the density function of $p(y^{(m_j)}_j|D, \vec{x},y_{j_s}^{*})$ can be computed using two different approximations as described below.

\vspace{1.0ex}
\noindent\textbf{Approximation 1 (MF-OSEMO-TG): }
As a consequence of {\bf Case II}, which states that $y^{(m_j)}_j\leq y_{j_s}^{*}$ also holds for all lower fidelities, the entropy of $p(y^{(m_j)}_j|D, \vec{x},y_{j_s}^{*})$ can also be approximated by the entropy of a truncated Gaussian distribution and expressed as follow:
\begin{align}
H(y^{(m_j)}_j|D, \vec{x},y^{(m_j)}_j\leq y_{j_s}^{*})\simeq &\frac{(1 + \ln(2\pi))}{2}+ \ln(\sigma^{(m_j)}_j(\vec{x})) +  \ln \Phi(\gamma_s^{(m_j)}(\vec{x})) \nonumber \\ &- \frac{\gamma_s^{(m_j)}(\vec{x}) \phi(\gamma_s^{(m_j)}(\vec{x}))}{2\Phi(\gamma_s^{(m_j)}(\vec{x}))}  \label{TGappriximation}
\end{align}
where $\gamma_s^{(m_j)}(\vec{x}) = \frac{y_{j_s}^{*} - \mu_j^{(m_j)}(\vec{x})}{\sigma_j^{(m_j)}(\vec{x})}$.\\ \\

\noindent\textbf{Approximation 2 (MF-OSEMO-NI): }
Although equation (\ref{TGappriximation}) is sufficient for computing the entropy for $m_j \neq M_j$, it can be improved by conditioning on a tighter inequality $y^{(M_j)}_j\leq y_{j_s}^{*}$ as compared to the general one, i.e., $y^{(m_j)}_j\leq y_{j_s}^{*}$.
As we show below, this improvement comes at the expense of not obtaining a final closed-form expression, but it can be efficiently computed via numerical integration. We apply the derivation of the entropy based on numerical integration for single-objective problem, proposed in \cite{takeno2019multi}, for the multi-objective setting. \\
Now, for calculating $H(y^{(m_j)}_j|D, \vec{x},y^{(m_j)}_j\leq y_{j_s}^{*})$ by replacing  $p(y^{(m_j)}_j|D, \vec{x},y^{(m_j)}_j \leq y_{j_s}^{*})$ with $p(y^{(m_j)}_j|D, \vec{x},y^{(M_j)}_j\leq y_{j_s}^{*})$ and using Bayes’ theorem, we have:
\begin{align}
    p(y^{(m_j)}_j|D, \vec{x},y^{(M_j)}_j\leq y_{j_s}^{*}) =\frac{ p(y^{(M_j)}_j\leq y_{j_s}^{*} | y^{(m_j)}_j,D,\vec{x}) p(y^{(m_j)}_j,D,\vec{x})}{p(y^{(M_j)}_j\leq y_{j_s}^{*}|D,\vec{x})}\label{totalprob}
\end{align}
Both the densities, $p(y^{(M_j)}_j\leq y_{j_s}^{*}|D,\vec{x})$ and $p(y^{(m_j)}_j,D,\vec{x})$ can be obtained from the predictive distribution of MF-GP model and is given as follows: 
\begin{align}
  & p(y^{(m_j)}_j,D,\vec{x})=\frac{\phi(\gamma_j^{(m_j)}(\vec{x}))}{\sigma_j^{(m_j)}} \label{prob1}\\
 &  p(y^{(M_j)}_j\leq y_{j_s}^{*}|D,\vec{x})=\Phi(\gamma_s^{(M_j)}(\vec{x}))) \label{prob2}
\end{align}
where $\gamma_j^{(m_j)}(\vec{x}) = \frac{y_j^{(m_j)} - \mu_j^{(m_j)}(\vec{x})}{\sigma_j^{(m_j)}(\vec{x})}$. Since MF-GP represents all fidelities as one unified Gaussian process, the joint marginal
distribution $p(y^{(M_j)}_j, y^{(m_j)}_j|D,\vec{x})$ can be immediately obtained from the posterior distribution of the corresponding model $\mathcal{GP}_j$ as given below:
\begin{align}
    p(y^{(M_j)}_j| y^{(m_j)}_j,\vec{x},D) \sim \mathcal{N}(\mu_j(\vec{x}),s_j^2(\vec{x}))
    \label{eqn:marginal_gaussian}
\end{align}
where $\mu_j(\vec{x})=\frac{\sigma_j^{2^{(m_jM_j)}}(\vec{x})(y_j^{(m_j)}-\mu_j^{m_j}(\vec{x}))}{\sigma_j^{2^{(m_j)}}(\vec{x})}$ and $s_j^2(\vec{x})=\sigma_j^{2^{(M_j)}}(\vec{x}) - \frac{(\sigma_j^{2^{(m_jM_j)}}(\vec{x}))^2}{\sigma_j^{2^{(m_j)}}(\vec{x})}$.
As a result, $p(y^{(M_j)}_j\leq y_{j_s}^{*} | y^{(m_j)}_j,D,\vec{x})$ is expressed as the cumulative distribution of the Gaussian in (\ref{eqn:marginal_gaussian}):
\begin{align}
    p(y^{(M_j)}_j\leq y_{j_s}^{*} | y^{(m_j)}_j,D,\vec{x})=\Phi(\frac{y_{j_s}^{*}-\mu_j(\vec{x})}{s_j(\vec{x})})\label{prob3}
\end{align}
By substituting (\ref{prob1}), (\ref{prob2}), and \ref{prob3} into (\ref{totalprob}) we get:
\begin{align}
     H(y^{(m_j)}_j|D, &\vec{x},y^{(M_j)}_j\leq y_{j_s}^{*})= - \int \Psi(y^{(m_j)}_j) \log(\Psi(y^{(m_j)}_j))dy^{(m_j)}_j \label{NIappriximation}
\end{align}
With $\Psi(y^{(m_j)}_j)= \Phi(\frac{y_{j_s}^{*}-\mu_j(\vec{x})}{s_j(\vec{x})}) \frac{\Phi(\gamma_s^{(M_j)}(\vec{x})))\phi(\gamma_j^{(m_j)}(\vec{x}))}{\sigma_j^{(m_j)}}$. Since this integral is over one-dimension variable $y^{(m_j)}_j$, numerical integration can result in a tight approximation. 

A complete description of the MF-OSEMO algorithm is given in Algorithm \ref{alg:OSEMO}. The blue colored steps correspond to computation of our
acquisition function via sampling.
\begin{algorithm*}[t]
\footnotesize
\caption{MF-OSEMO Algorithm}
\label{alg:OSEMO}
\textbf{Input}: input space $\mathfrak{X}$; $K$ blackbox objective functions where each function $f_j$ has multiple fidelities $M_j$ $\left(\{f_1^{(1)}(\vec{x}), \cdots, f_1^{(M_1)}(\vec{x})\},
\cdots,\{f_K^{(1)}(\vec{x}), \cdots, f_K^{(M_K)}(\vec{x})\}\right)$; and  total budget $\lambda_{Total}$
\begin{algorithmic}[1] 
\STATE Initialize multi-fidelity Gaussian process models $\mathcal{GP}_1, \cdots, \mathcal{GP}_K$ by evaluating at initial points $D$
\STATE \textbf{While} {$\lambda_{t} \leq \lambda_{total}$} \textbf{do}
 \STATE \quad for each sample $s \in {1,\cdots,S}$: 
 \STATE \quad \quad Sample highest-fidelity functions $\Tilde{f}_i^{(M_i)} \sim \mathcal{GP}_i, \quad \forall{i \in \{1,\cdots, K\}} $
 \STATE \quad \quad $\mathcal{Y}_s^{*} \leftarrow$ Pareto front of {\em cheap} multi-objective optimization over $(\Tilde{f}_1^{(M_1)}, \cdots, \Tilde{f}_K^{(M_K)})$
 \STATE \quad Find the next point to evaluate: select $(\vec{x}_{t},\vec{m}_t) \leftarrow \arg max_{\vec{x}\in \mathfrak{X},\vec{m}} \hspace{2 mm} \alpha_t(\vec{x},\vec{m},\mathcal{Y}^{*}) $
 \STATE \quad Update the total cost consumed: $\lambda_t \leftarrow \lambda_t + \lambda^{\vec{(m_t)}}$
\STATE \quad Aggregate data: $\mathcal{D} \leftarrow \mathcal{D} \cup \{(\vec{x}_{t}, \vec{y}_{t}^{\vec{m}})\}$ 
\STATE \quad Update models $\mathcal{GP}_1,\cdots, \mathcal{GP}_K$ 
\STATE \quad $t \leftarrow t+1$
\STATE \textbf{end while}
\STATE \textbf{return} Pareto front and Pareto set of $f_1(x), \cdots,f_K(x)$ based on $\mathcal{D}$\\
\STATE \textbf{Procedure} {$\alpha_t(\vec{x},\vec{m},\mathcal{Y}_s^{*})$}{}
\STATE  // Computes information gain (I)  about the posterior of true Pareto front $(\mathcal{Y}^*)$ per unit cost as a result of evaluating $\vec{x}$
\STATE // I = $H_1$ - $H_2$;\quad where $H_1$ = Entropy of $\vec{y}^{(\vec{m})}$ conditioned on $D$ and $\vec{x}$ \\
//\hspace{30mm} and $H_2$ = Expected entropy of $\vec{y}^{(\vec{m})}$ conditioned on $D$, $\vec{x}$ and $(\mathcal{Y}^*)$ 
\STATE Set $H_1 = H(\vec{y}^{(\vec{m})} \mid D, \vec{x}) = {K(1+\ln(2\pi))}/{2} +  \sum_{j = 1}^K  \ln (\sigma_j^{(m_j)}(\vec{x})) $ (entropy of K-factorizable Gaussian)
\STATE To compute $H_2  \simeq \frac{1}{S} \sum_{s = 1}^S \sum_{j=1}^K H(y^{(m_j)}_j|D, \vec{x},y_{j_s}^{*})$, initialize $H_2$ = 0
\FOR{each sample $\mathcal{Y}_s^*$}
\FOR{$j \in {1 \cdots K}$}
\STATE Set $y_{j_s}^{*}$ = maximum of $j$th component of all vectors in $\mathcal{Y}_s^*$
\STATE \textbf{If }{$m_j=M_j$} \quad // if evaluating $j$th function at highest fidelity
\STATE \quad  $H_2$ += $H(y^{(M_j)}_j|D, \vec{x},y^{(M_j)}_j \leq y_{j_s}^{*})$ (entropy of truncated Gaussian $p(y^{(M_j)}_j|D, \vec{x},\underline{y^{(M_j)}_j \leq y_{j_s}^{*}})$)
\STATE \textbf{ If }{$m_j\neq M_j$} \quad // if evaluating $j$th function at lower fidelity
\STATE \quad \quad // two approximations are provided
\STATE \quad \quad {\bf If} approximation = TG 
\STATE \quad \quad \quad  $H_2$ += $H(y^{(m_j)}_j|D, \vec{x},y^{(m_j)}_j\leq y_{j_s}^{*})$ (entropy of truncated Gaussian $p(y^{(M_j)}_j|D, \vec{x},\underline{y^{(m_j)}_j \leq y_{j_s}^{*}})$)
\STATE \quad \quad {\bf If} approximation = NI 
\STATE \quad \quad \quad  $H_2$ += $H(y^{(m_j)}_j|D, \vec{x},y^{(M_j)}_j\leq y_{j_s}^{*})$ (entropy computed via numerical integration)
\ENDFOR
\ENDFOR
\STATE  Divide by number of samples: $H_2$ = $H_2/S$
\STATE \textbf{return $(H_1 - H_2)/\lambda^{(\vec{m})}$} 
\end{algorithmic}
\end{algorithm*}

%% file: IMOCA_setup_Approach.tex
\newpage
\section{iMOCA Algorithm for Continuous-Fidelity MOO Problem}

In this section, we address the continuous-fidelity MOO problem, we have access to alternative functions through which we can evaluate cheaper approximations of objective functions by varying a continuous fidelity variable. To solve this problem, we propose an algorithm referred to as {\em {\bf i}nformation-Theoretic {\bf M}ulti-Objective Bayesian {\bf O}ptimization with {\bf C}ontinuous {\bf A}pproximations} (iMOCA). We first describe the complete details related to the continuous-fidelity MOO problem. Subsequently, we explain our proposed iMOCA algorithm with two mathematically different approximations of the output space entropy based acquisition function. 


\vspace{1.0ex}

\noindent {\bf Problem Setup (Continuous-Fidelity MOO Problem).} The continuous-fidelity MOO problem is the general version of the discrete multi-fidelity setting where we have access to $g_i(\vec{x},z_i)$ where $g_i$ is an alternative function through which we can evaluate cheaper approximations of $f_i$ by varying the fidelity variable $z_i \in \mathcal{Z}$ (continuous function approximations). Without loss of generality, let $\mathcal{Z}$=$\left[0, 1 \right]$ be the fidelity space. Fidelities for each function $f_i$ vary in the amount of resources consumed and the accuracy of evaluation, where $z_i$=0 and $z_i^*$=1 refer to the lowest and highest fidelity respectively. At the highest fidelity $z_i^*$, $g_i(\vec{x},z_i^{*})=f_i(\vec{x})$. 
The evaluation of an input $\vec{x}\in \mathcal{X}$ with fidelity vector $\vec{z} = [z_1, z_2, \cdots, z_K]$ produces an evaluation vector of $K$ values denoted by $\vec{y} \equiv [y_1, y_2,\cdots, y_K]$, where $y_i = g_i(\vec{x},z_i)$ for all $i \in \{1,2, \cdots, K\}$. Let $\mathcal{C}_i(\vec{x},z_i)$ be the cost of evaluating $g_i(\vec{x},z_i)$. Our goal is to approximate the optimal Pareto set $\mathcal{X}^*$ over the highest fidelities functions while minimizing the overall cost of function evaluations (experiments). For example, in rocket launching research, we need to find designs that trade-off return-time and angular distance using continuous-fidelity simulators (e.g., varying tolerance parameter to trade-off simulation time and accuracy) for design evaluations. Table~\ref{table:notations3} contains all the mathematical notations used in this section (iMOCA).
 
 \vspace{1.0ex}
 
\noindent{\bf Cost of Function Evaluations.} The total normalized cost of function evaluation is $\mathcal{C}(\vec{x},\vec{z}) = \sum_{i=1}^{K} \left( {\mathcal{C}_i(\vec{x},z_i)}/{\mathcal{C}_i(\vec{x},z_i^*)}\right)$. We normalize the cost of each function by the cost of its highest fidelity because the cost units of different objectives can be different.
 If the cost is known, it can be directly injected in the latter expression. However, in some real-world settings, the cost of a function evaluation can be only known after the function evaluation. In this case, the cost can be modeled by an independent Gaussian process. The predictive mean can be used during the optimization.  The final goal is to recover $\mathcal{X}^*$ while minimizing the total cost of function evaluations. 

\vspace{1.0ex}
 
\noindent{\bf Continuous-Fidelity GPs as Surrogate Models.} \label{surrogatesection}
Let $D$ = $\{(\vec{x}_i, \vec{y}_i,\vec{z}_i)\}_{i=1}^{t-1}$ be the training data from past $t$-1 function evaluations, where  $\vec{x}_i \in \mathcal{X}$ is an input and $\vec{y}_i = [y_1,y_2,\cdots,y_K]$ is the output vector resulting from evaluating functions $g_1, g_2,\cdots,g_K$ for $\vec{x}_i$ at fidelities $z_1, z_2,\cdots,z_K$ respectively. We learn from $\mathcal{D}$, $K$ surrogate statistical models $\mathcal{GP}_1,\cdots,\mathcal{GP}_K$, where each model $\mathcal{GP}_j$ corresponds to the $j$th function $g_j$. 
Continuous fidelity GPs (CF-GPs) are capable of modeling functions with continuous fidelities within a single model. Hence, we employ CF-GPs to build surrogate statistical models for each function. Specifically, we use the CF-GP model proposed in \cite{kandasamy2017multi}.
W.l.o.g, we assume that our functions $g_j$ are defined over the spaces $\mathcal{X}=[0,1]^d$ and $\mathcal{Z}=[0,1]$. Let $g_j \sim \mathcal{GP}_j(0,\kappa_j)$ such that $y_j=g_j(z_j,\vec{x})+\epsilon$, where $\epsilon \sim \mathcal{N}(0, \eta^2)$ and $\kappa: (\mathcal{Z} \times \mathcal{X})^2 \rightarrow \mathbb{R} $ is the prior covariance matrix defined on the product of input and fidelity spaces. 
\begin{align*}
    \kappa_j([z_j,\vec{x}],[z_j',\vec{x}'])= \kappa_{j\mathcal{X}}(\vec{x},\vec{x}') \cdot \kappa_{j\mathcal{Z}}(z_j,z_j')
\end{align*}
where $\kappa_{j\mathcal{X}},\kappa_{j\mathcal{Z}}$ are radial kernels over $\mathcal{X}$ and $\mathcal{Z}$ spaces respectively. $\mathcal{Z}$ controls the smoothness of $g_j$ over the fidelity space to be able to share information across different fidelities. 
A key advantage of this model is that it integrates all fidelities into one single GP for information sharing. 
We denote the posterior mean and standard deviation of $g_j$ conditioned on $D$ by $\mu_{g_j}(\vec{x},z_j)$ and $\sigma_{g_j}(\vec{x},z_j)$. We denote the posterior mean and standard deviation of the highest fidelity functions $f_j(\vec{x})=g_j(\vec{x},z_j^*)$ by $\mu_{f_j}(x)=\mu_{g_j}(\vec{x},z_j^*)$ and $\sigma_{f_j}(\vec{x})=\sigma_{g_j}(\vec{x},z_j^*)$ respectively. We define $\sigma_{g_j,f_j}^2(x)$ as the predictive co-variance between a lower fidelity $z_j$ and the highest fidelity $z_j^*$ at the same $\vec{x}$.
\begin{table*}[h!]
    \centering
    \resizebox{1\linewidth}{!}{
    \begin{tabular}{|c|c|}
    \hline
    {\bf Notation} & {\bf Definition} \\
    \hline \hline
              $g_1, g_2, \cdots, g_K$  & General objective functions with low and high fidelities\\
       \hline
$\Tilde{g}_j$ & Function sampled from the $j$th Gaussian process model at fidelity $z_j$\\
       \hline
       $z_1, z_2, \cdots, z_K$ & The fidelity variables for each function \\
       \hline

       $\vec{z}$ & Fidelities vector \\
       \hline
         $\vec{z}^*=[z_1^*, z_2^*, \cdots, z_K^*]$ & Fidelities vector with all fidelities at their highest value\\
       \hline
        $y_j$ &$j$th function $g_j$ evaluated at fidelity $z_j$  \\
       \hline
      $\vec{y}=[y_1,y_2,\cdots,y_K]$   & Output vector resulting from evaluating $g_1, g_2,\cdots,g_K$ \\ 
      & for $\vec{x}_i$ at fidelities $z_1, z_2,\cdots,z_K$ respectively\\  
      \hline 
            $\vec{f}=[f_1,f_2,\cdots,f_K]$   & Output vector resulting from evaluating functions $f_1, f_2,\cdots,f_K$ \\
            & or equivalently $g_1, g_2,\cdots,g_K$ for $\vec{x}_i$ at fidelities $z^*_1, z^*_2,\cdots,z^*_K$ respectively\\  
      \hline 
$\mathcal{C}_j(\vec{x},z_j)$ & Cost of evaluating $j$th function $g_j$ at fidelity $z_j$\\
\hline
$\mathcal{C}(\vec{x},\vec{z})$ & Total normalized cost $\mathcal{C}(\vec{x},\vec{z}) = \sum_{i=1}^{K} \left({\mathcal{C}_i(\vec{x},z_i)}/{\mathcal{C}_i(\vec{x},z_i^*)}\right)$  \\
\hline

$\mathcal{Z}$ & Fidelity space \\
\hline
$\mathcal{Z}_t^{(j)}$& Reduced fidelity space for function $g_j$ at iteration $t$ \\
\hline
$\mathcal{Z}_r$& Reduced fidelity space \\
\hline
$\xi$ & Information gap \\
\hline
$\beta_t^{(j)}$ &  Exploration/exploitation parameter for function $g_j$ at iteration $t$ \\
\hline

\hline

    \end{tabular}}
    \caption{Mathematical notations and their associated definition used in this section (iMOCA)}
    \label{table:notations3}
\end{table*}
\subsection{iMOCA Algorithm with Two Approximations}
We first describe the key idea behind our proposed iMOCA algorithm including the main challenges. Next, we present our  algorithmic solution to address those challenges.

\vspace{1.0ex}

\noindent {\bf Key Idea of iMOCA:} The acquisition function behind iMOCA employs principle of output space entropy search to select the sequence of input and fidelity-vector (one for each objective) pairs. iMOCA is applicable for solving MO problems in both continuous and discrete fidelity settings. {\em The key idea is to find the pair $\{\vec{x}_t, \vec{z}_t\}$ that maximizes the information gain $I$ per unit cost about the Pareto front of the highest fidelities} (denoted by $\mathcal{Y}^*$), where $\{\vec{x}_t, \vec{z}_t\}$ represents a candidate input $\vec{x}_t$ evaluated at a vector of fidelities $\vec{z}_t$ = $[z_1, z_2, \cdots, z_K]$ at iteration $t$. Importantly, iMOCA performs joint search over input space $\mathcal{X}$ and reduced fidelity space $\mathcal{Z}_r$ over fidelity vectors for this selection.
\begin{align}
    (\vec{x}_{t},\vec{z}_t) \leftarrow \arg max_{\vec{x}\in \mathcal{X},\vec{z}\in \mathcal{Z}_r}\hspace{2 mm}\alpha_t(\vec{x},\vec{z}) ~
        , \; \text{where} \quad \alpha_t(\vec{x},\vec{z}) &= I(\{\vec{x}, \vec{y},\vec{z}\}, \mathcal{Y}^* | D) / \mathcal{C}(\vec{x},\vec{z}) \label{af:def}
\end{align}
In the following sections, we describe the details and steps of our proposed algorithm iMOCA. We start by explaining the bottlenecks of continuous fidelity optimization due to the infinite size of the fidelity space followed 
by 
describing a principled approach to reduce the fidelity space. Subsequently, we present the computational steps of our proposed acquisition function: Information gain per unit cost for each candidate input and fidelity-vector pair. 

\subsubsection{Approach to Reduce Fidelity Search Space}\label{section-fidelity-reduction}

In this work, we focus primarily on MO problems with continuous fidelity space. The continuity of this space results in infinite number of fidelity choices. Thus, selecting an informative and meaningful fidelity becomes a major bottleneck. Therefore, we reduce the search space over fidelity-vector variables in a principled manner guided by the learned statistical models \cite{kandasamy2017multi}. Our fidelity space reduction method is inspired from BOCA for single-objective optimization \cite{kandasamy2017multi}. We apply the method in BOCA to each of the objective functions to be optimized in MO setting.

A favourable setting for continuous-fidelity methods would
be for the lower fidelities $g_j$ to be informative about the highest fidelity $f_j$. Let $h_j$ be the bandwith parameter of the fidelity kernel $\kappa_{j\mathcal{Z}}$ and let $\xi:\mathcal{Z}\rightarrow [0,1]$ be a measure of the gap in information about $g_j(.,z_j^*)$ when queried at $z_j\neq z_j^*$ with $\xi(z_j)\approx \frac{\Vert z_j - z_j^* \Vert}{h_j}$ for the squared exponential kernels \cite{kandasamy2017multi}. 
A larger $h_j$ will result in $g_j$ being smoother across $\mathcal{Z}$. Consequently, lower fidelities will be more informative about $f_j$ and the information gap $\xi(z_j)$ will be smaller.  

To determine an informative fidelity for each function in iteration $t$, we reduce the space $\mathcal{Z}$ and select $z_j$ from the subset $ \mathcal{Z}_t^{(j)}$ defined as follows:
\begin{align}
    \mathcal{Z}_t^{(j)}(\vec{x})=\{\{z_j \in  \mathcal{Z}_{\backslash \{z_j^*\}}, \sigma_{g_j}(\vec{x},z_j) > \gamma(z_j), \xi(z_j) > \beta_t^{(j)} \Vert \xi \Vert_{\infty} \} \cup \{z_j^*\}\}  \label{spacereduce}
\end{align} 
where $\gamma(z_j)=\xi(z_j) (\frac{\mathcal{C}_j(\vec{x},z_j)}{\mathcal{C}_j(\vec{x},z_j^*)})^q$ and $q=\frac{1}{p_j+d+2}$ with $p_j,d$ the dimensions of $\mathcal{Z}$ and $\mathcal{X}$ respectively. Without loss of generality, we assume that $p_j=1$. 
$\beta^{(j)}_t = \sqrt{0.5d \; \cdot \; log \; (2tl+1)}$ is the exploration/exploitation parameter \cite{kandasamy2017multi}. where, $l$ is the effective $L_1$ diameter of $\mathcal{X}$ and is computed by scaling each dimension by the inverse of the bandwidth of the SE kernel for that dimension.
We denote by $\mathcal{Z}_r=\{\mathcal{Z}_t^{(j)} , j \in \{1 \dots K\}\}$, the reduced fidelity space for all $K$ functions. 

We filter out the fidelities for each objective function at BO iteration $t$ using the above-mentioned two conditions. We provide intuitive explanation of these conditions below. 

\noindent \textbf{The first condition $\sigma_{g_j}(\vec{x},z_j) > \gamma(z_j)$:} A reasonable multi-fidelity strategy would query the cheaper fidelities in the beginning to learn about the function $g_j$ by consuming the least possible cost budget and later query from higher fidelities in order to gain more accurate information. Since the final goal is to optimize $f_j$, the algorithm should also query from the highest-fidelity. However, the algorithm might never query from higher fidelities due to their high cost. 
This condition will make sure that lower fidelities are likely to be queried, but not excessively and the algorithm will move toward querying higher fidelities as iterations progress. Since $\gamma(z_j)$ is monotonically
increasing in $\mathcal{C}_j$, this condition can be easily satisfied by cheap fidelities. However, if a fidelity is very far from $z_j^*$, then the information gap $\xi$ will increase and hence, uninformative fidelities would be discarded. Therefore, $\gamma(z_j)$ will guarantee achieving a good trade-off between resource cost and information.

\noindent \textbf{The second condition $\xi(z_j) > \beta_t^{(j)}\Vert \xi \Vert_{\infty}$:} We recall that if the first subset of $\mathcal{Z}_t^{(j)}$ is empty, the algorithm will automatically evaluate the highest-fidelity $z_j^*$. However, if it is not empty, and since the fidelity space is continuous (infinite number of choices for $z_j$), the algorithm might query fidelities that are very close to $z_j^*$ and would cost nearly the same as $z_j^*$ without being as informative as $z_j^*$. The goal of this condition is to prevent such situations by excluding fidelities in the small neighborhood of $z_j^*$ and querying $z_j^*$ instead. Since $\beta_t^{(j)}$ increases with $t$ and $\xi$ is increasing as we move away from $z_j^*$, this neighborhood is shrinking and the algorithm will eventually query $z_j^*$.

\subsubsection{Naive-CFMO: A Simple Continuous-Fidelity MO Baseline}\label{cf-naive-section}
In this section, we first describe a simple baseline approach referred to as {\em Naive-CFMO} to solve continuous-fidelity MO problems by combining the above-mentioned fidelity space reduction approach with existing multi-objective BO methods. Next, we explain the key drawbacks of Naive-CFMO and how our proposed iMOCA algorithm overcomes them.

A straightforward way to construct a continuous-fidelity MO method is to perform a two step selection process similar to the continuous-fidelity single-objective BO algorithm proposed in \cite{kandasamy2017multi}: 

\vspace{1.0ex}

\hspace{1.0ex} \noindent {\bf Step 1)} Select the input $\vec{x}$ that maximizes the acquisition function at the {\em highest fidelity}. This can be done using any existing multi-objective BO algorithm.

\vspace{1.0ex}

\hspace{1.0ex} \noindent {\bf Step 2)} Evaluate $\vec{x}$ at the {\em cheapest valid fidelity for each function} in the reduced fidelity space $ \mathcal{Z}_t^{(j)}(\vec{x})$ computed using the reduction approach mentioned in the previous section. Since we are studying information gain based methods in this work, we instantiate Naive-CFMO using the state-of-the-art information-theoretic MESMO algorithm \cite{belakaria2019max} for Step 1. Algorithm \ref{alg:CFMESMONaive} shows the complete pseudo-code of Naive-CFMO. 

\vspace{1.0ex}

\noindent {\bf Drawbacks of Naive-CFMO:} Unfortunately, Naive-CFMO has two major drawbacks.
\begin{itemize}
\setlength\itemsep{0em} 

\item The acquisition function solely relies on the highest-fidelity $f_j$. Therefore, it does not capture and leverage the statistical relation between  different fidelities and full-information provided by the global function $g_j$. 

\item Generally, there is a dependency between the fidelity space and the input space in continuous-fidelity problems. Therefore, selecting an input that maximizes the highest-fidelity and then evaluating it at a different fidelity can result in a mismatch in the evaluation process leading to poor performance and slower convergence.
\end{itemize}

\vspace{1.0ex}

\noindent {\bf iMOCA vs. Naive-CFMO:} Our proposed iMOCA algorithm overcomes the drawbacks of Naive-CFMO as follows. 

\begin{itemize}
\setlength\itemsep{0em} 

\item  iMOCA's acquisition function maximizes the information gain per unit cost across all fidelities by capturing the relation between fidelities and the impact of resource cost on information gain. 

\item iMOCA performs joint search over input and fidelity space to select the input variable $\vec{x} \in \mathcal{X}$ and fidelity variables $\vec{z} \in \mathcal{Z}_r$ while maximizing the proposed acquisition function. Indeed, our experimental results demonstrate the advantages of iMOCA over Naive-CFMO.

\end{itemize}

\subsubsection{Information-Theoretic Continuous-Fidelity Acquisition Function} 
In this section, we explain the technical details of the acquisition function behind iMOCA. We propose two approximations for the computation of information gain per unit cost.

The information gain in equation (\ref{af:def}) is defined as the expected reduction in entropy $H(.)$ of the posterior distribution $P(\mathcal{Y}^* | D)$ due to evaluating $\vec{x}$ at fidelity vector $\vec{z}$. Based on the symmetric property of information gain, the latter can be rewritten as follows:

\begin{align}
    I(\{\vec{x}, \vec{y},\vec{z}\}, \mathcal{Y}^{*} | D) &= H(\vec{y} | D, \vec{x},\vec{z}) - \mathbb{E}_{\mathcal{Y}^{*}} [H(\vec{y} | D,   \vec{x},\vec{z}, \mathcal{Y}^{*})]  \label{eqn_symmetric_ig3}
\end{align}
 In equation (\ref{eqn_symmetric_ig3}), the first term is the entropy of a $K$-dimensional Gaussian distribution  that can be computed in closed form as follows:

\begin{align}
H(\vec{y} | D, \vec{x},\vec{z}) = \sum_{j = 1}^K \ln (\sqrt{2\pi e} ~ \sigma_{g_j}(\vec{x},z_j))  \label{firstpart}
\end{align}

In equation (\ref{eqn_symmetric_ig3}), the second term is an expectation over the Pareto front of the highest fidelities $\mathcal{Y}^{*}$. This term can be approximated using Monte-Carlo sampling: 

\begin{align}
    \mathbb{E}_{\mathcal{Y}^{*}} [H(\vec{y} | D,   \vec{x},\vec{z}, \mathcal{Y}^{*})] \simeq \frac{1}{S} \sum_{s = 1}^S [H(\vec{y} | D,   \vec{x},\vec{z}, \mathcal{Y}^{*}_s)] \label{eqn_summation3}
\end{align}
where $S$ is the number of samples and $\mathcal{Y}^{*}_s$ denote a sample Pareto front obtained over the highest fidelity functions sampled  from $K$ surrogate models.  
To compute equation (\ref{eqn_summation3}), we provide algorithmic solutions to construct Pareto front samples $\mathcal{Y}^{*}_s$ and to compute the entropy with respect to a given Pareto front sample $\mathcal{Y}^{*}_s$. 

\vspace{1.0ex}

\noindent {\bf 1) Computing Pareto Front Samples:} We first sample highest fidelity functions \\$\Tilde{f}_1,\cdots,\Tilde{f}_K$ from the posterior CF-GP models via random Fourier features \cite{PES,random_fourier_features}. This is done similar to prior work \cite{PESMO,MES}. We solve a cheap MO optimization problem over the $K$ sampled functions $\Tilde{f}_1,\cdots,\Tilde{f}_K$ using the popular NSGA-II algorithm \cite{deb2002nsga} to compute the sample Pareto front $\mathcal{Y}^{*}_s$. 

\vspace{1.0ex}

\noindent {\bf 2) Entropy Computation for a Given Pareto Front Sample:} Let $\mathcal{Y}^{*}_s = \{\vec{v}^1, \cdots, \vec{v}^l \}$ be the sample Pareto front, where $l$ is the size of the Pareto front and each $\vec{v}^i = \{v_1^i,\cdots,v_K^i\}$ is a $K$-vector evaluated at the $K$ sampled highest-fidelity functions. The following inequality holds for each component $y_j$ of $\vec{y}$ = $(y_1, \cdots, y_K)$ in the entropy term $H(\vec{y} | D,   \vec{x},\vec{z}, \mathcal{Y}^{*}_s)$:
\begin{align}
 y_j &\leq f_s^{j*} \quad \forall j \in \{1,\cdots,K\} \label{inequality3}
\end{align}
where $f_s^{j*} = \max \{v_j^1, \cdots v_j^l \}$. Essentially, this inequality says that the $j^{th}$ component of $\vec{y}$ (i.e., $y_j$) is upper-bounded by a value, which is the maximum of $j^{th}$ components of all $l$ vectors $\{\vec{v}^1, \cdots, \vec{v}^l \}$ in the Pareto front $\mathcal{Y}^{*}_s$. Inequality (\ref{inequality3}) holds by the same proof of (\ref{inequality2}).
{\em For the ease of notation, we drop the dependency on $\vec{x}$ and $\vec{z}$. We use $f_j$ to denote $f_j(x)=g_j(x,z_j^*)$, the evaluation of the highest fidelity at $x$ and $y_j$ to denote $g_j(x,z_j)$ the evaluation of $g_j$ at a lower fidelity $z_j \neq z_j^*$.} 

By combining the inequality (\ref{inequality3}) and the fact that each function is modeled as an independent CF-GP, a common property of entropy measure allows us to decompose the entropy of a set of independent variables into a sum over entropies of individual variables \cite{information_theory}:
\begin{align}
H(\vec{y} | D,   \vec{x},\vec{z}, \mathcal{Y}^{*}_s) = \sum_{j=1}^K H(y_j|D, \vec{x},z_j,f_s^{j*}) \label{eqn_sep_ineq3}
\end{align} 
The computation of (\ref{eqn_sep_ineq3}) requires the computation of the entropy of $p(y_j|D, \vec{x},z_j,f_s^{j*})$. This is a conditional distribution that depends on the value of $z_j$ and can be expressed as $H(y_j|D, \vec{x},z_j,y_j\leq f_s^{j*})$. This entropy can be computed using two different approximations as described below.

\vspace{1.0ex}

\textbf{Truncated Gaussian Approximation (iMOCA-T): }
As a consequence of (\ref{inequality3}), which states that $y_j\leq f_s^{j*}$ also holds for all fidelities, the entropy of $p(y_j|D, \vec{x},z_j,f_s^{j*})$ can also be approximated by the entropy of a truncated Gaussian distribution and expressed as follows:
\begin{align}
&H(y_j|D, \vec{x},z_j,y_j\leq f_s^{j*})\simeq  \ln(\sqrt{2\pi e} ~ \sigma_{g_j}) +  \ln \Phi(\gamma_s^{(g_j)})- \frac{\gamma_s^{(g_j)} \phi(\gamma_s^{(g_j)})}{2\Phi(\gamma_s^{(g_j)})} ~ \text{where} ~ \gamma_s^{(g_j)} = \frac{f_s^{j*} - \mu_{g_j}}{\sigma_{g_j}}   \label{entropyapprox3}
\end{align}
From equations (\ref{eqn_summation3}), (\ref{firstpart}), and (\ref{entropyapprox3}), we get the final expression of iMOCA-T as shown below:
\begin{align}
    \alpha_t(\vec{x},\vec{z},\mathcal{Y}^{*})\simeq &\frac{1}{\mathcal{C}(\vec{x},\vec{z}) S}\sum_{j=1}^K \sum_{s=1}^S\left[  \frac{\gamma_s^{(g_j)}\phi(\gamma_s^{(g_j)})}{2\Phi(\gamma_s^{(g_j)})} - \ln(\Phi(\gamma_s^{(g_j)})) \right] \label{Tappriximation}
\end{align}

\vspace{1.0ex}

\textbf{Extended-skew Gaussian Approximation (iMOCA-E):} Although equation (\ref{Tappriximation}) is sufficient for computing the entropy, this entropy can be mathematically interpreted and computed with a different approximation. The condition $y_j \leq f_s^{j*}$, is originally expressed as $f_j \leq f_s^{j*}$. Substituting this condition with it's original equivalent, the entropy becomes $H(y_j|D, \vec{x},z_j,f_j\leq f_s^{j*}) $. Since $y_j$ is an evaluation of the function $g_j$ while $f_j$ is an evaluation of the function $f_j$, we observe that $y_j | f_j \leq f_s^{j*}$ can be approximated by an extended-skew Gaussian (ESG) distribution \cite{moss2020mumbo,azzalini1985class}. It has been shown that the differential entropy of an ESG does not have a closed form expression \cite{arellano2013shannon}. Therefore, we derive a simplified expression where most of the terms are analytical by manipulating the components of the entropy. We apply the derivation of the entropy based on ESG formulation, proposed by \citeA{moss2020mumbo}, to the multi-objective setting. 

In order to simplify the calculation $H(y_j|D, \vec{x},z_j,f_j\leq f_s^{j*})$,
let us define the normalized variable $\Gamma_{f_s^{j*}}$ as $\Gamma_{f_s^{j*}} \sim \frac{y_j - \mu_{g_j}}{\gamma_{g_j}} |f_j\leq f_s^{j*} $. $\Gamma_{f_s^{j*}}$ is distributed as an ESG with p.d.f whose mean $\mu_{\Gamma_{f_s^{j*}}}$ and variance $\sigma_{\Gamma_{f_s^{j*}}}$ are defined in Appendix \ref{full_derivation}. We define the predictive correlation between $y_j$ and $f_j$ as $\tau=\frac{\sigma_{g_j,f_j}^2}{\sigma_{g_j}\sigma_{f_j}}$. 
The entropy can be computed using the following expression. Due to lack of space, we only provide the final expression. Complete derivation for equations (\ref{entropyapprox2}) and (\ref{Eappriximation}) are provided in Appendix \ref{full_derivation}.
\begin{align}
    H(y_j|D, \vec{x},z_j,f_j\leq f_s^{j*}) &\simeq  \ln(\sqrt{2\pi e} ~\sigma_{g_j}) +\ln(\Phi(\gamma_s^{(f_j)})) - \tau^2\frac{\phi(\gamma_s^{(f_j)})\gamma_s^{(f_j)}}{2\Phi(\gamma_s^{(f_j)})}\nonumber \\
    & \quad - \mathbb{E}_{u \sim \Gamma_{f_s^{j*}}}\left[   \ln(\Phi(\frac{\gamma_s^{(f_j)}-\tau u}{\sqrt{1-\tau^2}}))\right]  \label{entropyapprox2}
\end{align}
From equations (\ref{eqn_summation3}), (\ref{firstpart}) and (\ref{entropyapprox2}), the final expression of iMOCA-E can be expressed as follow:
\begin{align}
    \alpha_t(\vec{x},\vec{z},\mathcal{Y}^{*})\simeq &\frac{1}{\mathcal{C}(\vec{x},\vec{z})S}\sum_{j=1}^K \sum_{s=1}^S \left[\tau^2\frac{\gamma_s^{(f_j)}\phi(\gamma_s^{(f_j)})}{2\Phi(\gamma_s^{(f_j)})} - \ln(\Phi(\gamma_s^{(f_j)})) +\mathbb{E}_{u \sim \Gamma_{f_s^{j*}}}\left[\ln(\Phi(\frac{\gamma_s^{(f_j)}-\tau u}{\sqrt{1-\tau^2}}))\right]\right]  \label{Eappriximation}
\end{align}
The expression given by equation (\ref{Eappriximation}) is mostly analytical except for the last term. We perform numerical integration via Simpson’s rule using $\mu_{\Gamma_{f_s^{j*}}} \mp \gamma \sqrt{\sigma(\Gamma_{f_s^{j*}})} $ as the integral limits. Since this integral is over one-dimension variable, numerical integration can result in a tight approximation with low computational cost. Complete pseudo-code of  iMOCA is shown in Algorithm \ref{alg:CFMESMO}.

\vspace{1.0ex}

\textbf{Generality of the Two Approximations:} We observe that for any fixed value of $\vec{x}$, when we choose the highest-fidelity for each function $\vec{z}$=$\vec{z}^*$, {\bf a)} For iMOCA-T, we will have $g_i=f_j$; and {\bf b)} For iMOCA-E, we will have $\tau = 1$. Consequently, both equation (\ref{Tappriximation}) and  equation (\ref{Eappriximation}) will degenerate to the acquisition function of MESMO optimizing only highest-fidelity functions given in equation (\ref{eqn_final}) in section \ref{Section-MESMO}.
 
\noindent The main advantages of our proposed acquisition function are: cost-efficiency, computational-efficiency, and robustness to the number of Monte-Carlo samples. Indeed, our experiments demonstrate these advantages over state-of-the-art single-fidelity MO algorithms.

\noindent
\begin{minipage}{0.49\textwidth}
\begin{algorithm}[H]
\centering
\scriptsize
\caption{iMOCA Algorithm}
\textbf{Input}: input space $\mathcal{X}$; $K$ blackbox functions $f_j$ and their continuous approximations $g_j$; total budget $\mathcal{C}_{total}$ 
\begin{algorithmic}[1] 
\STATE Initialize continuous fidelity Gaussian process $\mathcal{GP}_1, \cdots, \mathcal{GP}_K$ by initial points $D$
\STATE \textbf{While} {$\mathcal{C}_{t} \leq \mathcal{C}_{total}$} \textbf{do}
 \STATE \quad for each sample $s \in {1,\cdots,S}$: 
 \STATE \quad \quad Sample highest-fidelity functions $\Tilde{f}_j \sim \mathcal{GP}_j(.,z_j^*) $
 \STATE \quad \quad $\mathcal{Y}_s^{*} \leftarrow$ Solve {\em cheap} MOO over $(\Tilde{f}_1, \cdots, \Tilde{f}_K)$
 \STATE \quad Find the query based on $\mathcal{Y}^{*}=\{\mathcal{Y}_s^{*}, s \in \{1 \dots S\}\}$
\STATE \quad // Choose one of the two approximations
\STATE \quad {\bf If} approx = T // Use eq (\ref{Tappriximation}) for $\alpha_t$ (iMOCA-T)
\STATE \quad \quad select $(\vec{x}_{t},\vec{z}_t) \leftarrow \arg max_{\vec{x}\in \mathcal{X},\vec{z}\in \mathcal{Z}_r} \hspace{2 mm} \alpha_t(\vec{x},\vec{z},\mathcal{Y}^{*})$
\STATE \quad {\bf If} approx = E // Use eq (\ref{Eappriximation}) for $\alpha_t$ (iMOCA-E)
\STATE \quad\quad select $(\vec{x}_{t},\vec{z}_t) \leftarrow \arg max_{\vec{x}\in \mathcal{X},\vec{z}\in \mathcal{Z}_r} \hspace{2 mm}\alpha_t(\vec{x},\vec{z},\mathcal{Y}^{*})$ 
\STATE \quad Update the total cost: $\mathcal{C}_t \leftarrow \mathcal{C}_t + \mathcal{C}(\vec{x}_t,\vec{z}_t)$
\STATE \quad Aggregate data: $\mathcal{D} \leftarrow \mathcal{D} \cup \{(\vec{x}_{t}, \vec{y}_{t},\vec{z}_t)\}$ 
\STATE \quad Update models $\mathcal{GP}_1,\cdots, \mathcal{GP}_K$ 
\STATE \quad $t \leftarrow t+1$
\STATE \textbf{end while}
\STATE \textbf{return} Pareto front and Pareto set of black-box functions $f_1(x), \cdots,f_K(x)$
\end{algorithmic}
\label{alg:CFMESMO}
\end{algorithm}
\end{minipage}
\hfill
\begin{minipage}{0.49\textwidth}
\begin{algorithm}[H]
\centering
\scriptsize
\caption{Naive-CFMO Algorithm}
\textbf{Input}: input space $\mathcal{X}$; $K$ blackbox functions $f_j$ and their continuous approximations $g_j$; total budget $\mathcal{C}_{total}$
\begin{algorithmic}[1] 
\STATE Initialize continuous fidelity Gaussian process $\mathcal{GP}_1, \cdots, \mathcal{GP}_K$ by evaluating at initial points $D$
\STATE \textbf{While} {$\mathcal{C}_{t} \leq \mathcal{C}_{total}$} \textbf{do}
 \STATE \quad for each sample $s \in {1,\cdots,S}$: 
 \STATE \quad \quad Sample highest-fidelity functions $\Tilde{f}_j \sim \mathcal{GP}_j(.,z_j^*) $
 \STATE \quad \quad $\mathcal{Y}_s^{*} \leftarrow$ Solve {\em cheap} MOO over $(\Tilde{f}_1, \cdots, \Tilde{f}_K)$
 \STATE \quad Find the query based on $\mathcal{Y}^{*}=\{\mathcal{Y}_s^{*}, s \in \{1 \dots S\}\}$:
\STATE \quad // Use eq (\ref{eqn_final}) for $\alpha_t$ (MESMO)
 \STATE \quad  select $\vec{x}_{t} \leftarrow \arg max_{\vec{x}\in \mathcal{X}} \hspace{2 mm} \alpha_{t}(\vec{x},\mathcal{Y}^{*}) $
\STATE \quad \textbf{for } {$j \in {1 \cdots K}$} \textbf{do}
\STATE \quad \quad select $z_{j} \leftarrow  \arg min_{\vec{z_j}\in \mathcal{Z}_t^{(j)}(\vec{x_t}) \cup \{z_j^*\} } \hspace{2 mm} \mathcal{C}_i(x_t,z_j)  $ \vspace{0.9 mm}
\STATE \quad  Fidelity vector $\vec{z}_{t} \leftarrow [z_1 \dots z_K]$
 \STATE \quad Update the total cost: $\mathcal{C}_t \leftarrow \mathcal{C}_t + \mathcal{C}(\vec{x}_t,\vec{z}_t)$ 
\STATE \quad Aggregate data: $\mathcal{D} \leftarrow \mathcal{D} \cup \{(\vec{x}_{t}, \vec{y}_{t},\vec{z}_t)\}$
\STATE \quad Update models $\mathcal{GP}_1,\cdots, \mathcal{GP}_K$ 
\STATE \quad $t \leftarrow t+1$
\STATE \textbf{end while}
\STATE \textbf{return} Pareto front and Pareto set of black-box functions $f_1(x), \cdots,f_K(x)$
\end{algorithmic}
\label{alg:CFMESMONaive}
\end{algorithm}
\end{minipage}

%% file: Experiments.tex
\section{Experiments and Results}

In this section, we first describe the experimental evaluation of MESMO (single-fidelity algorithm), MF-OSEMO (discrete multi-fidelity algorithm) and iMOCA (continuous-fidelity algorithm) on synthetic and real-world engineering problems. Subsequently, we present experimental results of MESMOC (constrained MO algorithm) on two real-world engineering problems, namely, electrified aviation power system design and analog circuit design.

\subsection{Experimental Evaluation of iMOCA, MF-OSEMO, and MESMO}

We mainly present the results for iMOCA with MESMO and MF-OSEMO as baselines for the following reasons: First, iMOCA is the generalisation of both MESMO and MF-OSEMO to the most general setting (continuous-fidelity); and  second, the performance, robustness, and effectiveness of MESMO and MF-OSEMO have been shown in \cite{belakaria2019max} and \cite{belakaria2020multi} respectively.

\vspace{1.0ex}

\noindent {\bf Experimental Setup.} In our experiments, we employed CF-GP models as described in section \ref{surrogatesection} with squared exponential kernels. We initialize the surrogate models of all functions with the same number of points selected randomly from both lower and higher fidelities.  We compare iMOCA with several baselines: six state-of-the-art single-fidelity MO algorithms (ParEGO, SMSego, EHI, SUR, PESMO, and MESMO) and one naive continuous-fidelity baseline that we proposed in Section \ref{cf-naive-section}. 
 We employ the code for ParEGO, PESMO, SMSego, EHI, and SUR from the BO library Spearmint\footnote{github.com/HIPS/Spearmint/tree/PESM}. The code for all four of our algorithms are available in public Github repositories.
 We provide more details about the algorithms parameters, libraries, and computational resources in the Appendix \ref{addtional_info}.
For experiments in discrete fidelity setting, the number of fidelities is very limited. Thus, the fidelity space reduction method deem meaningless in this case. Therefore, we employ iMOCA without fidelity space reduction for those scenarios. Additionally, we  compare to the state-of-the-art discrete fidelity method MF-OSEMO. MF-OSEMO has two variants: MF-OSEMO-TG and MF-OSEMO-NI. Since MF-OSEMO-TG has the same formulation as iMOCA-T and provide similar results, we compare only to MF-OSEMO-NI. 

\subsubsection{Synthetic Benchmarks}
\begin{figure*}[h!] 
    \centering
    \begin{minipage}{1\textwidth}
    \centering
    \begin{minipage}{0.49\textwidth}
        \centering
        \includegraphics[width=0.89\textwidth]{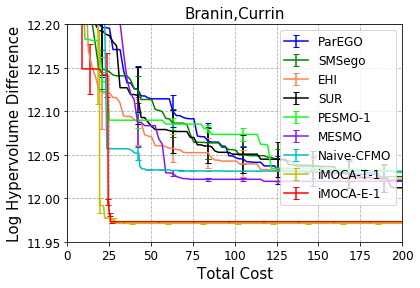} 
    \end{minipage}\hfill
    \begin{minipage}{0.49\textwidth}
        \centering
        \includegraphics[width=0.86\textwidth]{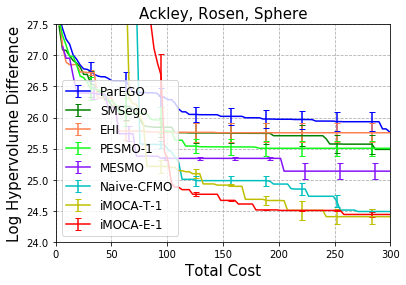} 
    \end{minipage}
    \end{minipage} %
    \begin{minipage}{1\textwidth}
            \centering
    \begin{minipage}{0.49\textwidth}
        \centering
        \includegraphics[width=0.85\textwidth]{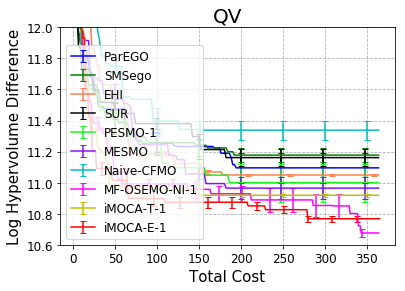} 
    \end{minipage}\hfill
    \begin{minipage}{0.49\textwidth}
        \centering
        \includegraphics[width=0.84\textwidth]{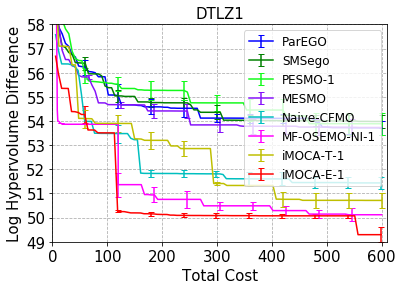} 
    \end{minipage}
    \end{minipage}
          \begin{minipage}{1\textwidth}
            \centering
    \begin{minipage}{0.49\textwidth}
        \centering
        \includegraphics[width=0.83\textwidth]{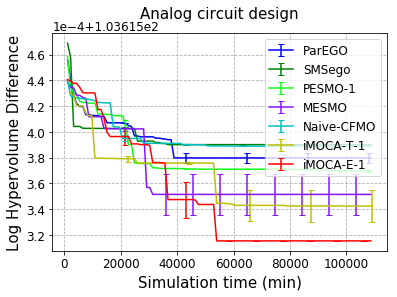} 
    \end{minipage}\hfill
    \begin{minipage}{0.49\textwidth}
        \centering
        \includegraphics[width=0.84\textwidth]{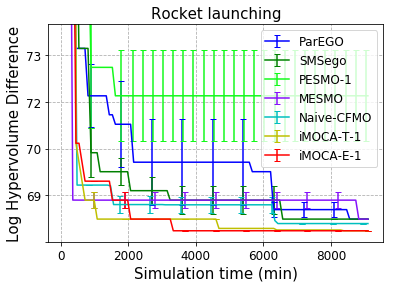} 
    \end{minipage}
    \end{minipage} %
      \begin{minipage}{1\textwidth}
    \centering

        \begin{minipage}{0.49\textwidth}
        \centering
        \includegraphics[width=0.86\textwidth]{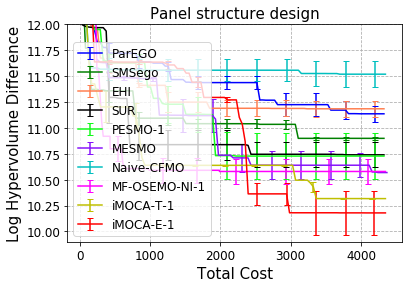} 
    \end{minipage}
    \hfill
    \begin{minipage}{0.49\textwidth}
        \centering
        \includegraphics[width=0.83\textwidth]{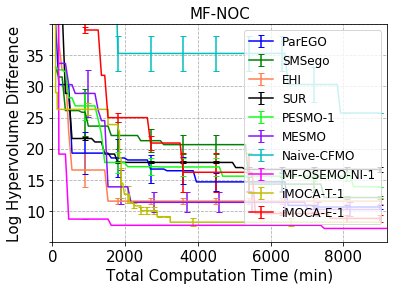} 
    \end{minipage}
    \end{minipage}
\caption{Results of iMOCA and the baselines algorithms on synthetic benchmarks and real-world problems. The $PHV$ metric is presented against the total resource cost of function evaluations.} 
\label{syntheticexp}
\end{figure*}

\begin{figure*}[h!] 
    \centering
    \begin{minipage}{1\textwidth}
    \centering
    \begin{minipage}{0.49\textwidth}
        \centering
        \includegraphics[width=0.89\textwidth]{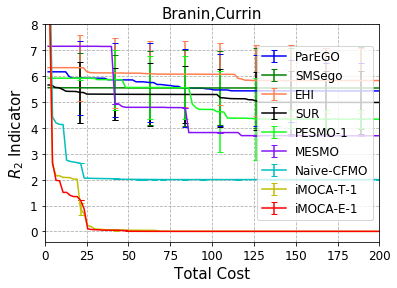} 
    \end{minipage}\hfill
    \begin{minipage}{0.49\textwidth}
        \centering
        \includegraphics[width=0.89\textwidth]{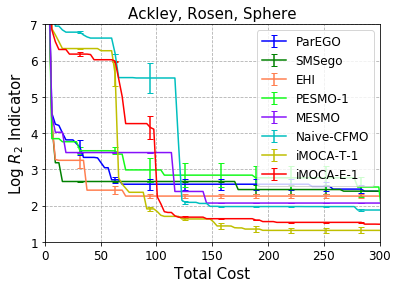} 
    \end{minipage}
    \end{minipage} %
    \begin{minipage}{1\textwidth}
            \centering
                \begin{minipage}{0.49\textwidth}
        \centering
        \includegraphics[width=0.85\textwidth]{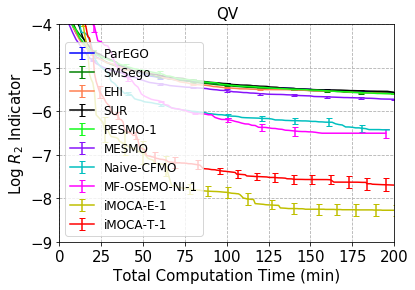} 
    \end{minipage}\hfill
    \begin{minipage}{0.49\textwidth}
        \centering
        \includegraphics[width=0.85\textwidth]{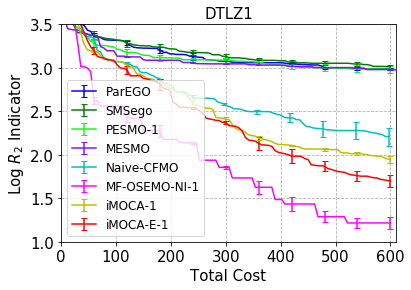} 
    \end{minipage}

    \end{minipage}
          \begin{minipage}{1\textwidth}
            \centering
    \begin{minipage}{0.49\textwidth}
        \centering
        \includegraphics[width=0.85\textwidth]{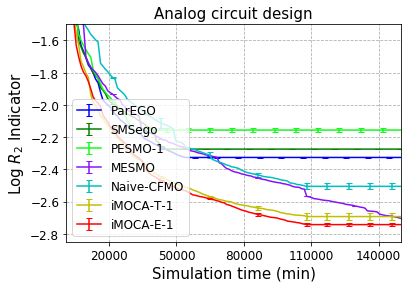} 
    \end{minipage}\hfill
    \begin{minipage}{0.49\textwidth}
        \centering
        \includegraphics[width=0.83\textwidth]{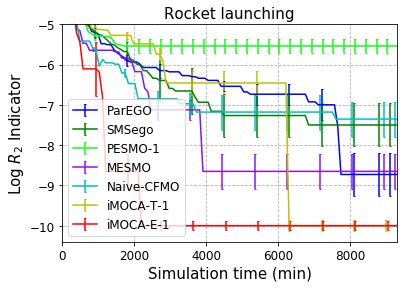} 
    \end{minipage}
    \end{minipage}
      \begin{minipage}{1\textwidth}
    \centering
    \begin{minipage}{0.49\textwidth}
        \centering
        \includegraphics[width=0.85\textwidth]{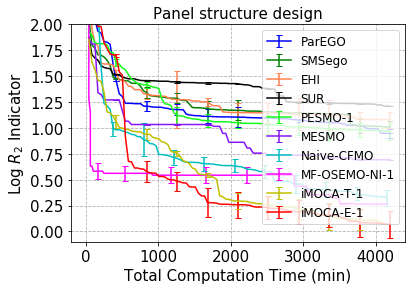} 
    \end{minipage}\hfill
    \begin{minipage}{0.49\textwidth}
        \centering
        \includegraphics[width=0.83\textwidth]{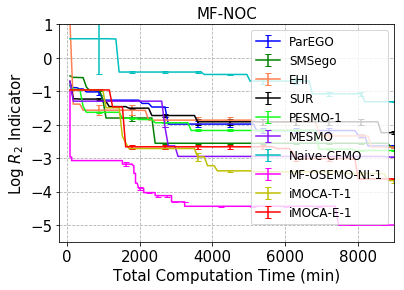} 
    \end{minipage}
    \end{minipage}
    
\caption{Results of iMOCA and the baselines algorithms on synthetic benchmarks and real-world problems. The $R_2$ metric is presented against the total resource cost of function evaluations.}
\label{syntheticexpR2}
\end{figure*}

We evaluate our most general algorithm iMOCA and baselines on four different synthetic benchmarks.  We construct two problems using a combination of 
benchmark functions for continuous-fidelity and single-objective optimization \cite{simulationlib}: \emph{Branin,Currin} (with $K$=2, $d$=2) and \emph{Ackley, Rosen, Sphere} (with $K$=3, $d$=5). To show the effectiveness of iMOCA on settings with discrete fidelities, we employ two of the known general MO benchmarks: \emph{QV} (with $K$=2, $d$=8) and \emph{DTLZ1} (with $K$=6, $d$=5) \cite{habib2019multiple,shu2018line}. We provide their complete details in Appendix \ref{sec:appSynthetic}. The titles of plots in Fig. \ref{syntheticexp}, Fig. \ref{syntheticexpsample}, and Fig. \ref{syntheticexpR2} denote the corresponding experiments. 

\subsubsection{Real-world Engineering Design Optimization Problems} 
We evaluate iMOCA and baselines on four real-world design optimization problems from diverse engineering domains. We provide the details of these problems below. 

\vspace{1.0ex}

\noindent {\bf 1) Analog Circuit Design Optimization.} Design of a 
voltage regulator via Cadence circuit simulator that imitate the real hardware \cite{DATE-2020,hong2019dual}.  The simulation time can be adjusted to vary the simulation from fast and inaccurate to long and accurate. Each candidate circuit design is defined by 33 input variables ($d$=33). We optimize nine objectives: efficiency, four output voltages, and four output ripples. This problem has a continuous-fidelity space with cost varying from 10 mins to 120 mins. 

\vspace{1.0ex}

\noindent {\bf 2) Panel Structure Design for Large Vessels.} The deck structure in large vessels commonly require the design of panels  
resisting uni-axial compression in the direction of the stiffeners \cite{zhu2014multi}. We consider optimizing the trade-off between two objective functions: weight and strength of the panel. These functions depend on six input variables ($d$=6): one of them is the number of stiffeners used and five others relating to the plate thickness and stiffener dimensions. This problem has a discrete fidelity setting: two fidelities with computational costs 1 min and 21 mins respectively.  

\vspace{1.0ex}

\noindent {\bf 3) Rocket Launching Simulation.} Rocket launching studies \cite{hasbun2012classical} require several long and computationally-expensive simulations to reach an optimal design. In this problem, we have three input variables ($d=3$): mass of fuel, launch height, and launch angle. 
The three objective functions are return time, angular distance, and difference between the launch angle and the radius at the point of launch. The simulator has a parameter that can be adjusted to perform continuous fidelity simulations.
We employ the parameter range to vary the cost from 0.05 to 30 mins.

\vspace{1.0ex}

\noindent {\bf 4) Network-On-Chip Design.} Communication infrastructure is critical for efficient data movement in hardware chips \cite{NOC1,NOC2,NOC3,NOC4} and they are designed using cycle-accurate simulators. We consider a dataset of 1024 configurations of a network-on-chip with ten input variables ($d$=10) \cite{rodinia-benchmark}. 
We optimize two objectives: latency and energy. This problem has two discrete fidelities with costs 3 mins and 45 mins respectively.


\subsubsection{Results and Discussion}
We compare iMOCA with both approximations (iMOCA-T and iMOCA-E) to all baselines. We employ two known metrics for evaluating the quality of a given Pareto front: {\em Pareto hypervolume ($PHV$)} metric and $R_2$ indicator. 
$PHV$ \cite{zitzler1999evolutionary} is defined as the volume between a reference point and the given Pareto front; and $R_2$ \cite{picheny2013benchmark} is a distance-based metric defined as the average distance between two Pareto-fronts.
We report both the difference in the hyper-volume, and the average distance between an optimal Pareto front $(\mathcal{F^{*}}) $ and the best recovered Pareto front estimated by optimizing the posterior mean of the models at the highest fidelities \cite{PESMO}. The mean and variance of $PHV$ and $R_2$ metrics across 10 different runs are reported as a function of the total cost.

Fig. \ref{syntheticexp} shows the $PHV$ results of all the baselines and iMOCA for synthetic and real-world experiments (Fig. \ref{syntheticexpR2} shows the corresponding $R_2$ results). We observe that iMOCA consistently outperforms all baselines. Both iMOCA-T and iMOCA-E have lower converge cost. Additionally, iMOCA-E shows a better convergence rate than iMOCA-T. This result can be explained by its tighter approximation. Nevertheless, iMOCA-T displays very close or sometimes better results than iMOCA-E. This demonstrates that even with loose approximation, using the iMOCA-T approximation can provide consistently competitive results using less computation time. For experiments with the discrete fidelity setting, iMOCA most of the times outperformed MF-OSEMO or produced very close results. It is important to note that MF-OSEMO is an algorithm designed specifically for the discrete-fidelity setting. Therefore, the competitive performance of iMOCA shows its effectiveness and generalisability.

Figure \ref{syntheticexpsample} in appendix \ref{addtional_res} shows the results of evaluating iMOCA and PESMO with varying number of Monte-Carlo samples $S \in \{1,10,100\}$. For ease of comparison and readability, we present these results in two different figures side by side. We observe that the convergence rate of PESMO is dramatically affected by the number of MC samples $S$. However, iMOCA-T and iMOCA-E maintain a better performance consistently even with a single sample. These results strongly demonstrate that our method {\em iMOCA is much more robust to the number of Monte-Carlo samples}.

\begin{table}[h!]
\centering
\caption{
{\em Best} convergence cost from all baselines $\mathcal{C}_B$, {\em Worst} convergence cost for iMOCA $\mathcal{C}$, and cost reduction factor $\mathcal{G}$.
}\label{tab:costreduction}
\resizebox{0.5\linewidth}{!}{
\begin{tabular}{lllll}  
\toprule
Name & BC  & ARS   & Circuit  & Rocket   \\
\midrule
$\mathcal{C}_B$ & 200  & 300   & 115000 & 9500  \\
\midrule
$\mathcal{C}$ & 30  & 100  & 55000 & 2000  \\
\midrule
$\mathcal{G}$ & 85\%  & 66.6\%  & 52.1\% & 78.9\% \\
\bottomrule 
\end{tabular}
}
\end{table}
\noindent {\bf Cost Reduction Factor.} 
We also provide the \textit{cost reduction factor} for experiments with continuous fidelities, which is the percentage of gain in the convergence cost when compared to the best performing baseline (the earliest cost for which any of the single-fidelity baselines converge). Although this metric gives advantage to baselines, the results in Table \ref{tab:costreduction} show a consistently high gain ranging from $52.1\%$ to $85\%$. 

\subsection{Experimental Evaluation of MESMOC}
\textbf{Experimental Setup:}
In this section, we compare MESMOC with PESMOC \cite{garrido2019predictive}, the state-of-the-art BO algorithm for solving constrained MO problems and MESMOC+ \cite{fernandez2020max}, the concurrent approach which also relies on the same principle of output space entropy search. Due to lack of BO approaches for constrained MO setting, we also compare to known genetic algorithms: NSGA-II \cite{deb2002nsga} and MOEAD \cite{zhang2007moea}. However, they require large number of function evaluations to converge which is not practical for the optimization of expensive functions. We employ a GP based statistical model with squared exponential (SE) kernel in all our experiments. The hyper-parameters are estimated after every five function evaluations (iterations). We initialize the GP models for all functions by sampling the initial points at random.
We employ the code for PESMOC and MESMOC+ from the BO library Spearmint\footnote{github.com/EduardoGarrido90/Spearmint}. We employ NSGA-II and MOEAD from the Platypus library\footnote{platypus.readthedocs.io/en/latest/getting-started.html\#installing-platypus}. Our code for MESMOC is available at the following Github repository \footnote{github.com/belakaria/MESMOC}. We provide additional details about the algorithms parameters, libraries, and computational resources in the Appendix \ref{addtional_info}.

\subsubsection{Real-world Engineering Design Problems}

Below we provide the details of the two real-world problems and associated optimization task that are employed for our experimental evaluation.

\vspace{1.0ex}

\noindent {\bf 1) Electrified Aviation Power System Design.}  We consider optimizing the design of electrified aviation power system of unmanned aerial vehicle (UAV) via a time-based static simulation. The UAV system architecture consists of a central Li-ion
battery pack, hex-bridge DC-AC inverters, PMSM motors, and necessary wiring \cite{belakaria2020machine}. Each candidate input consists of a set of $5$ ($d$=5) variable design parameters such as the battery pack configuration (battery cells in series, battery cells in parallel) and motor
size (number of motors, motor stator winding length, motor stator winding turns). We minimize two objective functions: mass and total energy. This problem has $5$ black-box constraints: 
\begin{align}
    &{C}_0: \text{Maximum final depth of discharge} \leq 75 \% \nonumber \\
    &{C}_1: \text{Minimum cell voltage}\geq 3 V \nonumber\\
    &{C}_2: \text{Maximum motor temperature} \leq \ang{125} C \nonumber\\
    &{C}_3:\text{Maximum inverter temperature} \leq \ang{120} C \nonumber\\
    & {C}_5:\text{ Maximum modulation index} \leq 1.3 \nonumber
\end{align}

\vspace{1.0ex}

\noindent {\bf 2) Analog Circuit Optimization Domain.} We consider optimizing the design of a multi-output switched-capacitor voltage regulator via Cadence circuit simulator that imitates the real hardware \cite{DATE-2020}. This circuit relies on a dynamic frequency switching clock. Each candidate circuit design is defined by 33 input variables ($d$=33). The first 24 variables are the width, length, and unit of the eight capacitors of the circuit $W_i,L_i,M_i ~ \forall i \in 1\cdots 8$. The remaining input variables are four output voltage references $V_{ref_i}~ \forall i \in 1\cdots 4$ and four resistances $R_i ~ \forall i \in 1\cdots 4$ and a switching frequency $f$. We optimize nine objectives: maximize efficiency $Eff$, maximize four output voltages $V_{o_1} \cdots V_{o_4}$, and minimize four output ripples $OR_{1} \cdots OR_{4}$.
Our problem has a total of nine constraints. Since some of the constraints have upper bounds and lower bounds, they are defined in the problem by 15 different constraints:
\begin{align}
    &{C}_0: Cp_{total} \simeq 20 nF  ~ with ~ Cp_{total}= \sum_{i=1}^8 (1.955W_iL_i+0.54(W_i+L_i))M_i \nonumber \\
    &{C}_1~ to~ {C}_4:  V_{o_i}\geq V_{ref_i} ~ \forall \in{1\cdots4} \nonumber \\
    &{C}_5\ to\ {C}_{8\ }:\ \ OR_{lb}\le OR_i\le OR_{ub} ~ \forall i\ \in{1\cdots4} \nonumber\\
    &{C}_9:Eff \le 100\% \nonumber
\end{align}
where $OR_{lb}$ and $OR_{ub}$ are the predefined lower-bound and upper-bound of $OR_i$ respectively. $Cp_{total}$ is the total capacitance of the circuit.

\vspace{1.5ex}

\subsubsection{Results and Discussion}
\begin{figure}[h!] 
    \centering
    \begin{minipage}{0.49\textwidth}
        \centering
        \includegraphics[width=0.89\textwidth]{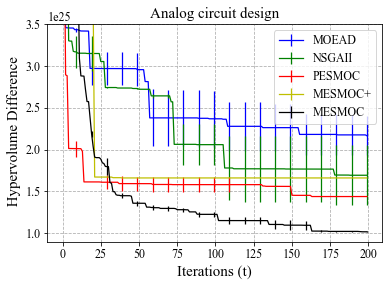} 
    \end{minipage}\hfill
    \begin{minipage}{0.49\textwidth}
        \centering
        \includegraphics[width=0.89\textwidth]{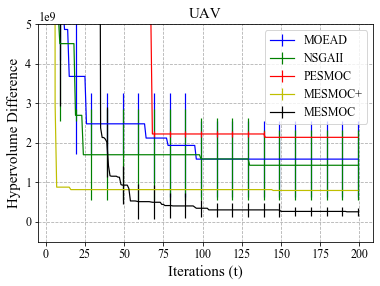} 
    \end{minipage}
\caption{Results of different constrained multi-objective algorithms including MESMOC. The $PHV$ metric is shown as a function of the number function evaluations.}\label{fig:hv}
    \end{figure}

We evaluate the performance of our algorithm and the baselines using the Pareto hypervolume (PHV) metric. PHV is a commonly employed metric to measure the quality of a given Pareto front \cite{zitzler1999evolutionary}. 
Figure \ref{fig:hv} shows that MESMOC outperforms existing baselines. It recovers a better Pareto front with a significant gain in the number of function evaluations.
Both of these experiments are motivated by real-wold engineering applications where further analysis of the designs in the Pareto front is crucial. 

\vspace{1.0ex}

\noindent {\bf Electrified Aviation Power System Design.}
In this setting, the input space is discrete with 250,000 combinations of design parameters. Out of the entire design space, only 9\% of design combinations passed all the constraints and only five points are in the optimal Pareto front. From a domain expert perspective, satisfying all the constraints is critical. Hence, the results reported for the hypervolume include only points that satisfy all the constraints. Despite the hardness of the problem, 90\% (180 out of 200 inputs) of the designs selected by MESMOC satisfy all the constraints while for MESMOC+, PESMOC, MOEAD, and NSGA-II, this was 49\% (98 out of 200), 1.5\% (3 out of 200 inputs),  9.5\% (19 out of 200 inputs), and 7.5\% (15 out of 200 inputs) respectively. MESMOC was not able to recover all the five points from the optimal Pareto front. However, it was able to closely approximate the optimal Pareto front and recover better designs than the baselines.

\vspace{1.0ex}

\noindent {\bf Analog Circuit Design Optimization.} In this setting, the input space is continuous, consequently there is an infinite number of candidate designs. From a domain expert perspective, satisfying all the constraints is not critical and is impossible to achieve. The main goal is to satisfy most of the constraints (and getting close to satisfying the threshold for violated constraints) while reaching the best possible objective values. Therefore, the results reported for the hypervolume include all the evaluated points. In this experiment, the efficiency of circuit is the most important objective function. The table in Figure 2 shows the optimized circuit parameters from different algorithms.

    \begin{figure}[h!]
  \centering
  \includegraphics[width=.8\linewidth]{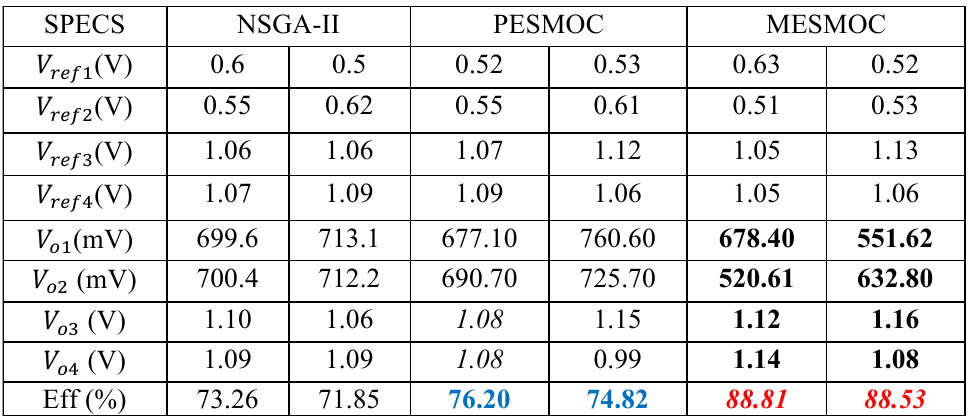}

\caption{Comparison table of optimized circuit parameters obtained from different algorithms (designs are selected from the Pareto set prioritized by efficiency)}  \label{fig:table}
\end{figure}

All algorithms can generate design parameters for the circuit that meets the voltage reference requirements. The optimized circuit using MESMOC can achieve the highest conversion efficiency of 88.81\% (12.61\% improvement when compared with PESMOC with fixed frequency optimization and 17.86\% improvement when compared with NSGA-II) with similar output ripples.  The circuit with optimized parameters can generate the target output voltages within the range of 0.52V to 0.76V (1/3x ratio) and 0.99V to 1.17V (2/3x ratio) under the loads varying from 14 Ohms to 1697 Ohms. 

%% file: appendix.tex
\section{Full Derivation of iMOCA's Acquisition Function}\label{full_derivation}

 Our goal is to derive a full approximation for iMOCA algorithm. In this appendix, we provide the technical details of the extended-skew Gaussian approximation (iMOCA-E) for the computation of the information gain per unit cost.

 The information gain in equation (\ref{af:def}) is defined as the expected reduction in entropy $H(.)$ of the posterior distribution $P(\mathcal{Y}^* | D)$ due to evaluating $\vec{x}$ at fidelity vector $\vec{z}$. Based on the symmetric property of information gain, we can rewrite it as shown below:

\begin{align}
    I(\{\vec{x}, \vec{y},\vec{z}\}, \mathcal{Y}^{*} | D) &= H(\vec{y} | D, \vec{x},\vec{z}) - \mathbb{E}_{\mathcal{Y}^{*}} [H(\vec{y} | D,   \vec{x},\vec{z}, \mathcal{Y}^{*})]  \label{eqn_symmetric_igA}
\end{align}
 In equation (\ref{eqn_symmetric_igA}), the first term is the entropy of a $K$-dimensional Gaussian distribution  that can be computed in closed form as follows:
\begin{align}
H(\vec{y} | D, \vec{x},\vec{z}) = \sum_{j = 1}^K \ln (\sqrt{2\pi e} ~ \sigma_{g_j}(\vec{x},z_j))  \label{firstpartA}
\end{align}
The second term of equation (\ref{eqn_symmetric_igA}) is an expectation over the Pareto front of the highest fidelities $\mathcal{Y}^{*}$. This term can be approximated using Monte-Carlo sampling: 
\begin{align}
    \mathbb{E}_{\mathcal{Y}^{*}} [H(\vec{y} | D,   \vec{x},\vec{z}, \mathcal{Y}^{*})] \simeq \frac{1}{S} \sum_{s = 1}^S [H(\vec{y} | D,   \vec{x},\vec{z}, \mathcal{Y}^{*}_s)] \label{eqn_summationA}
\end{align}
In the main paper, we showed that :
\begin{align}
 y_j &\leq f_s^{j*} \quad \forall j \in \{1,\cdots,K\} \label{inequalityA}
\end{align}
By combining the inequality (\ref{inequalityA}) and the fact that each function is modeled as an independent CF-GP, a common property of entropy measure allows us to decompose the entropy of a set of independent variables into a sum over entropies of individual variables \cite{information_theory}:
\begin{align}
H(\vec{y} | D,   \vec{x},\vec{z}, \mathcal{Y}^{*}_s) \simeq \sum_{j=1}^K H(y_j|D, \vec{x},z_j,f_s^{j*}) \label{eqn_sep_ineqA}
\end{align} 
In what follows, we provide details of iMOCA-E approximation to compute $H(y_j|D, \vec{x},z_j,f_s^{j*})$.

The condition $y_j \leq f_s^{j*}$, is originally expressed as $f_j \leq f_s^{j*}$. Substituting this condition with it's original equivalent, the entropy becomes $H(y_j|D, \vec{x},z_j,f_j\leq f_s^{j*}) $. Since $y_j$ is an evaluation of the function $g_j$ and $f_j$ is an evaluation of the function $f_j$, we make the observation that $y_j | f_j \leq f_s^{j*}$ can be approximated by an extended-skew Gaussian (ESG) distribution \cite{azzalini1985class}. It had been shown that the differential entropy of an ESG does not have a closed-form expression \cite{arellano2013shannon}. Therefore, we derive a simplified expression where most of the terms are analytical by manipulating the components of the entropy as shown below. 

In order to simplify the calculation $H(y_j|D, \vec{x},z_j,f_j\leq f_s^{j*})$, we start by deriving an expression for its probability distribution. Based on the definition of the conditional distribution of a bi-variate normal, $f_j | y_j$ is normally distributed with mean $\mu_{f_j}+ \frac{\sigma_{f_j}}{\sigma_{g_j}}\tau(y_j-\mu_{g_j})$ and variance $\sigma_{f_j}^2(1-\tau)^2$, where $\tau= \frac{\sigma_{g_j,f_j}^2}{\sigma_{g_j}\sigma_{f_j}}$ is the predictive correlation between $y_j$ and $f_j$. We can now write the cumulative distribution function for $y_j|f_j\leq f_s^{j*}$ as shown below:
\begin{align*}
    &P(y_j\leq u | f_j \leq f_s^{j*}) = \frac{P(y_j\leq u , f_j \leq f_s^{j*})}{P( f_j \leq f_s^{j*})} = \frac{\int_{-\infty}^u \phi \left(\frac{\theta - \mu_{g_j}}{\sigma_{g_j}}\right) \Phi \left( \frac{f_s^{j*}-\mu_{f_j}- \frac{\sigma_{f_j}}{\sigma_{g_j}}\tau(\theta-\mu_{g_j})}{\sqrt{\sigma_{f_j}^2(1-\tau)^2}} \right) d\theta}{\sigma_{g_j} \Phi \left(\frac{f_s^{j*}-\mu_{f_j}}{\sigma_{f_j}}\right)}
\end{align*}
Let us define the normalized variable $\Gamma_{f_s^{j*}}$ as $\Gamma_{f_s^{j*}}\sim \frac{y_j - \mu_{g_j}}{\gamma_{g_j}} |f_j\leq f_s^{j*} $. After differentiating with respect to $u$, we can express the probability density function for $\Gamma_{f_s^{j*}}$ as: 
\begin{align*}
    P(u)=\frac{\phi(u)}{\Phi(\gamma_s^{(f_j)})}\Phi(\frac{\gamma_s^{(f_j)}-\tau u}{\sqrt{1-\tau^2}})
\end{align*}
which is the density of an ESG with mean and variance defined as follows: 
\begin{align}
   & \mu_{\Gamma_{f_s^{j*}}}=\tau \frac{\phi(\gamma_s^{(f_j)})}{\Phi(\gamma_s^{(f_j)})},  \sigma_{\Gamma_{f_s^{j*}}}=1- \tau^2\frac{\phi(\gamma_s^{(f_j)})}{\Phi(\gamma_s^{(f_j)})}\left[\gamma_s^{(f_j)}+\frac{\phi(\gamma_s^{(f_j)})}{\Phi(\gamma_s^{(f_j)})} \right]\label{momentsesg}
\end{align}
Therefore, we can express the entropy of the ESG as shown below:
\begin{align}
    H(\Gamma_{f_s^{j*}})= - \int P(u) \ln(P(u))du
\end{align}
We also derive a more simplified expression of the iMOCA-E acquisition function based on ESG.  For a fixed sample $f_s{^j*}$, $H(\Gamma_{f_s^{j*}})$ can be decomposed as follows: 
\begin{align}
    H(\Gamma_{f_s^{j*}})&=\mathbb{E}_{u \sim \Gamma_{f_s^{j*}}}\left[ -\ln(\phi(u)) + \ln(\Phi(\gamma_s^{(f_j)})) -\ln(\Phi(\frac{\gamma_s^{(f_j)}-\tau u}{\sqrt{1-\tau^2}}))\right]  \label{eq1A}
\end{align}
We expand the first term as shown below:
\begin{align}
    \mathbb{E}_{u \sim \Gamma_{f_s^{j*}}}\left[ -\ln(\phi(u))\right]=\frac{1}{2} \ln(2\pi) +\frac{1}{2}  \mathbb{E}_{u \sim \Gamma_{f_s^{j*}}}\left[ u^2\right]
\end{align}
From the mean and variance of $\Gamma_{f_s^{j*}}$ in equation (\ref{momentsesg}), we get: 
\begin{align}
    \mathbb{E}_{u \sim \Gamma_{f_s^{j*}}}\left[ u^2\right] &= \mu_{\Gamma_{f_s^{j*}}}^2 +  \sigma_{\Gamma_{f_s^{j*}}}=1- \tau^2\frac{\phi(\gamma_s^{(f_j)})\gamma_s^{(f_j)}}{\Phi(\gamma_s^{(f_j)})}
\end{align}
We note that the final entropy can be computed using the following expression.
\begin{align}
    H(y_j|D, \vec{x},z_j,y_j\leq f_s^{j*})=  H(\Gamma_{f_s^{j*}})+ \ln(\sigma_{g_j}) \label{ESappriximationA}
\end{align}
By combining equations (\ref{eq1A}) and (\ref{ESappriximationA}), we get:
\begin{align}
    H(y_j|D, \vec{x},z_j,f_j\leq f_s^{j*}) &\simeq \ln(\sqrt{2\pi e} ~\sigma_{g_j}) +\ln(\Phi(\gamma_s^{(f_j)})) - \tau^2\frac{\phi(\gamma_s^{(f_j)})\gamma_s^{(f_j)}}{2\Phi(\gamma_s^{(f_j)})}\nonumber \\
    & \quad - \mathbb{E}_{u \sim \Gamma_{f_s^{j*}}}\left[   \ln(\Phi(\frac{\gamma_s^{(f_j)}-\tau u}{\sqrt{1-\tau^2}}))\right]  \label{entropyapprox2app}
\end{align}

From equations (\ref{eqn_summationA}), (\ref{firstpartA}), and (\ref{entropyapprox2app}), the final expression of iMOCA-E can be expressed as follows:
\begin{align*}
    \alpha_t(\vec{x},\vec{z},\mathcal{Y}^{*})\simeq &\frac{1}{\mathcal{C}(\vec{x},\vec{z})S}\sum_{j=1}^K \sum_{s=1}^S \tau^2\frac{\gamma_s^{(f_j)}\phi(\gamma_s^{(f_j)})}{2\Phi(\gamma_s^{(f_j)})} - \ln(\Phi(\gamma_s^{(f_j)})) +\mathbb{E}_{u \sim \Gamma_{f_s^{j*}}}[\ln(\Phi(\frac{\gamma_s^{(f_j)}-\tau u}{\sqrt{1-\tau^2}}))] 
\end{align*}
Since the differential entropy of an ESG cannot be computed analytically, we
perform numerical integration via Simpson’s rule using $\mu_{\Gamma_{f_s^{j*}}} \mp \gamma \sqrt{\sigma_{\Gamma_{f_s^{j*}}}} $ as the integral limits. In practice, we set $\gamma$ to 5. Since this integral is over one-dimension variable, numerical integration can result in a tight approximation with small amount of computation.

\section{Additional Experiments and Results}

\subsection{Description of Synthetic Benchmarks}
\label{sec:appSynthetic}

In what follows, we provide complete details of the synthetic benchmarks employed in this paper. Since our algorithm is designed for maximization settings, we provide the benchmarks in their maximization form. 

\subsubsection*{1) Branin, Currin experiment}

In this experiment, we construct a multi-objective problem using a combination of existing single-objective optimization benchmarks \cite{kandasamy2017multi}. It has two functions with two dimensions ($K$=2 and $d$=2). 

\textbf{Branin Function:}
We use the following function where $\mathcal{C}(z) = 0.05 + z^{6.5}$ 

$$ g(\vec{x},z) = - \left(a(x_2 - b(z)x_1^2 + c(z)x_1 - r)^2 + s(1-t(z))cos(x_1) + s \right)$$

where $a = 1$, $b(z)=5.1/(4\pi^2) - 0.01(1-z)$,
$c(z) = 5/\pi - 0.1(1-z)$, $r=6$, $s=10$ and $t(z)=1/(8\pi) + 0.05(1-z)$.

\textbf{Currin Exponential Function:}
We use $\mathcal{C}(z) = 0.1 + z^2$
\begin{align*}
g(\vec{x},z) &=- \left(1-0.1(1-z)\exp\left(\frac{-1}{2x_2}\right)\right)
  \left(\frac{2300x_1^3 + 1900x_1^2 + 2092x_1 + 60}{100x_1^3 + 
  500x_1^2 + 4x_1 + 20}\right).
\end{align*}

\subsubsection*{2) Ackley, Rosen, Sphere  experiment}

In this experiment, we construct a multi-objective problem using a combination of existing single-objective optimization benchmarks \cite{wu2018continuous}. It has three functions with five dimensions ($K$=3 and $d$=5). For all functions, we employed $\mathcal{C}(z) = 0.05 + z^{6.5}$ 

\textbf{Ackley Function} 

$$ g(\vec{x},z)= -\left(-20 \exp \left[-0.2{\sqrt {\frac{1}{d}\sum_{i=1}^d x_i^{2}}}\right] -\exp \left[\frac{1}{d}\sum_{i=1}^d \cos (2\pi x_i)\right]+e+20\right)-0.01(1-z)$$

\textbf{Rosenbrock Function:}

$$ g(\vec{x},z)=-\sum _{i=1}^{d-1}\left[100\left(x_{i+1}-x_{i}^{2}+0.01(1-z)\right)^{2}+\left(1-x_{i}\right)^{2}\right]$$

\textbf{Sphere Function:}
$$g(\vec{x},z)=-\sum _{i=1}^{d}x_{i}^{2}-0.01(1-z)$$

\subsubsection*{3) DTLZ1 experiment}

In this experiment, we solve a problem from the general multi-objective optimization benchmarks \cite{habib2019multiple}. We have six functions with five dimensions ($K$=6 and $d$=5) with a discrete fidelity setting. Each function has three fidelities in which $z$ takes three values from $\{0.2,0.6,1\}$ with $z^*$=1. The cost of evaluating each fidelity function is $\mathcal{C}(z)$=$\{0.01,0.1,1\}$

$$g_j(\vec{x},z)=f_j(\vec{x})-e(\vec{x},z)$$

$f_1(\vec{x})=-(1+r)0.5\Pi_{i=1}^5 x_i$

$f_j(\vec{x})=-(1+r)0.5(1-x_{6-j+1})\Pi_{i=1}^{6-j} x_i$ with $j= 2 \dots 5$

$f_6(\vec{x})=-(1+r)0.5(1-x_{1})$

$r=100[d+\sum_{i=1}^d((x_i-0.5)^2)-cos(10 \pi (x_i-0.5))]$

$e(\vec{x},z)=\sum_{i=1}^d \alpha(z)cos(10 \pi \alpha(z)x_i+0.5 \pi \alpha(z)+\pi)$ 
 with $\alpha(z)=1-z$ 
 
\subsubsection*{4) QV experiment}

In this experiment, we solve a problem from the general multi-objective optimization benchmarks \cite{shu2018line}. We have two functions with eight dimensions ($K$=2 and $d$=8) with a discrete fidelity setting.

\textbf{Function 1} has only one fidelity which is the highest fidelity
$$ f_1(\vec{x})=-(\frac{1}{d}\sum_{i=1}^d(x_i^2-20 \pi x_i+10))^{\frac{1}{4}}  $$

\textbf{Function 2} has two fidelities with cost $\{0.1,1\}$ respectively and the following expressions: 

High fidelity: $ f_2(\vec{x}, High)=-(\frac{1}{d}\sum_{i=1}^d((x_i-1.5)^2-20 \pi (x_i-1.5)+10))^{\frac{1}{4}}  $

Low fidelity: $f_2(\vec{x}, Low)=-(\frac{1}{d}((\sum_{i=1}^d(\vec{\alpha[i]}(x_i-1.5)^2-20 \pi (x_i-1.5)+10))^{\frac{1}{4}}  $

with $\vec{\alpha}$=$[0.9,1.1,0.9,1.1,0.9,1.1,0.9,1.1]$

\subsection{Additional Information About Experimental Setup}\label{addtional_info}
\noindent
\textbf{Experimental Setup For Our Proposed Algorithms:}
\begin{itemize}
    \item The hyper-parameters are estimated after every five function evaluations (BO iterations) for MESMO and MESMOC. For iMOCA and MF-OSEMO, the number of evaluations would be higher due to the low cost of lower fidelities. Therefore, the hyper-parameters are estimated every twenty iterations.
    \item During the computation of Pareto front samples, we solve a cheap MO optimization problem over sampled functions using NSGA-II. We use  Platypus\footnote{platypus.readthedocs.io/en/latest/getting-started.html\#installing-platypus} library for the implementation. For NSGA-II, the most important parameter is the number of function calls. We experimented with several values. We noticed that increasing this number does not result in any performance improvement for our algorithms. Therefore, we fixed it to 1500 for all our experiments.
\end{itemize}
\textbf{Parameters Used for NSAG-II and MOEAD as Constrained Baselines:}
\begin{itemize}
    \item Since we allow only 200 evaluations for MESMOC and PESMOC, we also set the number of functions evaluations for NSGA-II and MOEAD to 200. We leave any other parameter to the default value provided by the Platypus library.
\end{itemize}
\textbf{Computational Resources}
\begin{itemize}
    \item We performed all experiments on a machine with the following configuration: Intel i7-7700K CPU @ 4.20GHz with 8 cores and 32 GB memory.
\end{itemize}

\subsection{Additional Results}\label{addtional_res}

\begin{figure*}[h!] 
    \centering
    \begin{minipage}{1\textwidth}
    \centering
    \begin{minipage}{0.49\textwidth}
        \centering
        \includegraphics[width=0.89\textwidth]{imoca_figures/bc_phv_1_new_new_L.png} 
    \end{minipage}\hfill
    \begin{minipage}{0.49\textwidth}
        \centering
        \includegraphics[width=0.89\textwidth]{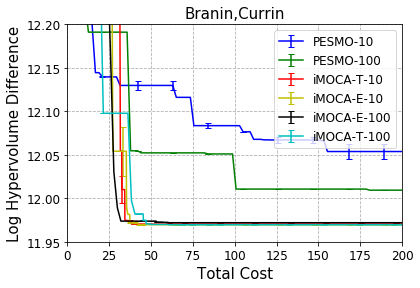} 
    \end{minipage}
    \end{minipage} %
    \begin{minipage}{1\textwidth}
            \centering
    \begin{minipage}{0.49\textwidth}
        \centering
        \includegraphics[width=0.85\textwidth]{imoca_figures/ARS_phv1_new_new_L.png} 
    \end{minipage}\hfill
    \begin{minipage}{0.49\textwidth}
        \centering
        \includegraphics[width=0.85\textwidth]{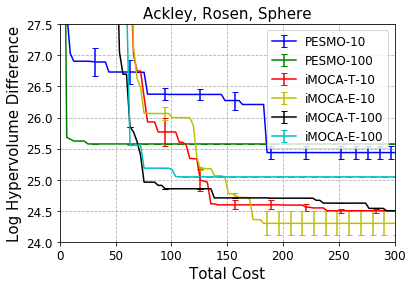} 
    \end{minipage}
    \end{minipage}
        \caption{Results of synthetic benchmarks showing the effect of varying the number of Monte-Carlo samples for iMOCA, MESMO, and PESMO. The hypervolume difference is shown against the total resource cost of function evaluations.}
\label{syntheticexpsample}
\end{figure*}